\documentclass[a4paper, 10pt, conference, australian]{ieeeconf}      

\IEEEoverridecommandlockouts                              
\usepackage{mathptmx} 
\usepackage{times} 

\title{\LARGE\bf
Path Signatures for Diversity in Probabilistic Trajectory Optimisation
}

\author{Lucas Barcelos$^{*,1}$, Tin Lai$^{1}$, Rafael Oliveira$^{2}$, Paulo Borges$^{2}$ and Fabio Ramos$^{1,3}$
    \thanks{$^{*}${\tt\small lucas.barcelos@sydney.edu.au}}%
    \thanks{$^{1}$The University of Sydney, Australia}%
    \thanks{$^{2}$CSIRO, Australia}%
    \thanks{$^{3}$NVIDIA, United States}%
}

\usepackage[utf8]{inputenc}
\usepackage[T1]{fontenc}
\usepackage{babel}  
\usepackage{microtype}  
\usepackage{bm}
\usepackage{amsfonts}
\usepackage[bb=dsserif]{mathalpha}  

\usepackage{mathtools}  
\usepackage{amssymb}
\newtheorem{definition}{Definition}
\usepackage{xfrac}  

\usepackage{pseudocode}
\usepackage[linesnumbered, vlined, ruled]{algorithm2e}

\usepackage{float}  
\usepackage[colorlinks,bookmarksopen,bookmarksnumbered,citecolor=red,urlcolor=red]{hyperref}
\usepackage{xurl}  
\usepackage{graphicx}  
\usepackage{subcaption}  
\usepackage{booktabs, longtable}  
\usepackage{csquotes}  
\usepackage[capitalise]{cleveref}
\usepackage{siunitx}
\usepackage{ifthen}
\usepackage{balance}
\usepackage{lipsum}

\usepackage[yyyymmdd]{datetime}  

\usepackage[
backend=biber,
style=ieee,
mincitenames=1,
maxcitenames=3,
giveninits=true,
url=false,
natbib=true,
]{biblatex}
\usepackage{xparse}
\usepackage{mleftright}
\mleftright{}  

\DeclarePairedDelimiterX{\parens}[1]{\lparen}{\rparen}{%
  \renewcommand{\given}{\;\delimsize\vert\nonscript\;\mathopen{}}
  #1
}  
\DeclarePairedDelimiter{\bracks}{\lbrack}{\rbrack}

\DeclarePairedDelimiterX{\set}[1]\{\}{%
  \renewcommand\given{\SetDelim[\delimsize]}
  #1
}
\NewDocumentCommand{\pargs}{ooo}{%
  \IfBooleanTF{#1}%
    {\parens*{#2}}%
    {\IfValueTF{#3}%
      {\parens[#2]{#3}}%
      {\parens{#2}}%
    }
}

\NewDocumentCommand{\bargs}{ooo}{%
  \IfBooleanTF{#1}%
    {\bracks*{#2}}%
    {\IfValueTF{#3}%
      {\bracks[#2]{#3}}%
      {\bracks{#2}}%
    }
}

\NewDocumentCommand{\fn}{oooooo}{%
  #2
  \IfValueT{#3}{_{#3}}\IfValueT{#4}{^{#4}}
  \IfValueT{#5}{\pargs[#1][#5][#6]}
}

\NewDocumentCommand{\op}{oooooo}{%
  #2
  \IfValueT{#3}{_{#3}}\IfValueT{#4}{^{#4}}
  \IfValueT{#5}{\bargs[#1][#5][#6]}
}

\def\mclimits_#1{\limits_{\mathclap{#1}}}  

\newcommand*{\given}{\mid}
\newcommand*{\field}[1]{\mathbb{\MakeUppercase{#1}}}  
\newcommand*{\setSymbol}[1]{{\mathcal{\MakeUppercase{#1}}}}  
\DeclarePairedDelimiterX{\norm}[1]{\lVert}{\rVert}{#1}  
\DeclarePairedDelimiterX{\inner}[2]{\langle}{\rangle}{#1, #2}  

\newcommand*{\R}{\field{R}} 

\newcommand*{\argmax}{\operatornamewithlimits{\arg\max}}
\newcommand*{\argmin}{\operatornamewithlimits{\arg\min}}

\newcommand*{\seq}[1]{\MakeUppercase{#1}}

\newcommand*{\dif}{{\operatorname{d}}}
\NewDocumentCommand{\grad}{e{_^}}{%
  \mathop{}\!
  \nabla%
  \IfValueT{#1}{_{\!#1}}
  \IfValueT{#2}{^{#2}}
}

\newcommand\Bbbone{%
  \ifdefined\mathbbb%
    \mathbbb{1}%
  \else%
    \boldsymbol{\mathbb{1}}%
  \fi}

\newcommand*{\llim}{a}  
\newcommand*{\ulim}{b}  

\newcommand*{\noise}{\rv{\eta}}  

\newcommand*{\anyscalar}{s}

\newcommand*{\anyvector}{\vec{x}}
\newcommand*{\othervector}{\vec{y}}
\newcommand*{\anypath}{\seq{x}}
\newcommand*{\otherpath}{\seq{y}}
\newcommand*{\anyerror}{\vec{e}}
\NewDocumentCommand{\anyfunction}{se{_^}oo}{\fn[#1][f][#2][#3][#4][#5]}
\NewDocumentCommand{\otherfunction}{se{_^}oo}{\fn[#1][g][#2][#3][#4][#5]}
\NewDocumentCommand{\randomfunction}{se{_^}oo}{\fn[#1][\tilde{\anyfunction}][#2][#3][#4][#5]}

\NewDocumentCommand{\map}{se{_^}oo}{\fn[#1][\operator{M}][#2][#3][#4][#5]}
 
\renewcommand{\vec}[1]{{\boldsymbol{\mathbf{#1}}}}
\newcommand*{\mat}[1]{{\MakeUppercase{#1}}}
\newcommand*{\transpose}{\mathsf{T}}

\NewDocumentCommand{\prob}{se{_^}oo}{\fn[#1][\mathbb{P}][#2][#3][#4][#5]}

\newcommand*{\measure}[1]{\mathbb{\MakeUppercase{#1}}}

\NewDocumentCommand{\pdf}{sme{_^}oo}{\fn[#1][#2][#3][#4][#5][#6]}
\NewDocumentCommand{\pPdf}{se{_^}oo}{\fn[#1][p][#2][#3][#4][#5]}
\NewDocumentCommand{\expectation}{se{_^}oo}{\op[#1][\mathbb{E}][#2][#3][#4][#5]}

\newcommand*{\kl}[2]{D_\mathrm{KL} \bigl(#1||#2\bigr)}

\NewDocumentCommand{\normal}{se{_^}oo}{\fn[#1][\mathcal{N}][#2][#3][#4][#5]}  
\newcommand*{\dirac}[1]{\delta\lparen#1\rparen}

\newcommand*{\rv}[1]{\MakeUppercase{#1}}  

\newcommand*{\qClass}{\setSymbol{Q}}
\NewDocumentCommand{\qDistribution}{se{_^}oo}{\fn[#1][\measure{Q}][#2][#3][#4][#5]}
\NewDocumentCommand{\qPdf}{se{_^}oo}{\fn[#1][q][#2][#3][#4][#5]}
\NewDocumentCommand{\pDistribution}{se{_^}oo}{\fn[#1][\measure{p}][#2][#3][#4][#5]}
\NewDocumentCommand{\fDensity}{se{_^}oo}{\fn[#1][\hat{f}][#2][#3][#4][#5]}

\NewDocumentCommand{\signature}{se{_^}oo}{\fn[#1][\operatorname{S}][#2][#3][#4][#5]}
\newcommand*{\sigDegree}{d}
\newcommand*{\sigLength}{l}
\newcommand*{\sigChannels}{c}
\newcommand*{\sigBatch}{n}
\NewDocumentCommand{\colPdf}{se{_^}oo}{\fn[#1][p_{\text{col}}][#2][#3][#4][#5]}
\NewDocumentCommand{\selfPdf}{se{_^}oo}{\fn[#1][p_{\text{self}}][#2][#3][#4][#5]}

\newcommand*{\splineKnots}{N_{k}}

\newcommand*{\primIdx}{i}
\newcommand*{\secIdx}{j}  
\newcommand*{\thirdIdx}{k}
\newcommand*{\tIdx}{t}

\newcommand*{\nSamples}{N_{s}}

\newcommand*{\state}{\vec{x}}
\newcommand*{\stateSpace}{\setSymbol{X}}
\newcommand*{\stateSeq}{\seq{x}}

\newcommand*{\stateInit}{\state_{s}}
\newcommand*{\stateGoal}{\state_{g}}

\newcommand*{\trajSpace}{\setSymbol{P}_{\stateSpace}}

\newcommand*{\ctrl}{\vec{u}}
\newcommand*{\ctrlSpace}{\setSymbol{U}}
\newcommand*{\ctrlSeq}{\seq{u}}

\newcommand*{\dynCost}{\costFn_{\text{dyn}}}
\newcommand*{\lenCost}{\costFn_{\text{len}}}
\newcommand*{\colCost}{\costFn_{\text{col}}}
\newcommand*{\scolCost}{\costFn_{\text{s-col}}}
\NewDocumentCommand{\fcol}{se{_^}oo}{\fn[#1][f_{\text{col}}][#2][#3][#4][#5]}
\NewDocumentCommand{\fscol}{se{_^}oo}{\fn[#1][f_{\text{s-col}}][#2][#3][#4][#5]}
\newcommand*{\costWeight}{\vec{w}}
\NewDocumentCommand{\costFn}{se{_^}oo}{\fn[#1][\mathcal{C}][#2][#3][#4][#5]}

\newcommand*{\optimality}{\mathcal{O}}

\NewDocumentCommand{\transFn}{se{_^}oo}{\fn[#1][\vec{f}][#2][#3][#4][#5]}

\newcommand*{\ctrlHorizon}{H}

\newcommand*{\temperature}{\lambda}

\newcommand*{\polSamples}{N_{\operatorname{a}}}

\newcommand*{\ctrlAuth}{\mat{\Sigma}}

\newcommand*{\wState}{\vec{x}}
\newcommand{\wSpace}{\mathcal{W}}
\newcommand{\wDim}{w}
\newcommand*{\cState}{\vec{q}}
\newcommand{\cSpace}{\mathcal{Q}}
\newcommand{\cDim}{d}
\newcommand*{\bodyPts}{b}
\newcommand*{\interPts}{p}
\NewDocumentCommand{\ffk}{se{_^}oo}{\fn[#1][\psi][#2][#3][#4][#5]}
\NewDocumentCommand{\jacob}{se{_^}oo}{\fn[#1][\vec{J}][#2][#3][#4][#5]}

\newcommand*{\partSize}{N_{p}}
\newcommand*{\partDim}{p}
\NewDocumentCommand{\scoreFunc}{se{_^}oo}{\fn[#1][\vec{\phi}][#2][#3][#4][#5]}
\newcommand*{\stepSize}{\epsilon}

\NewDocumentCommand{\logLik}{se{_^}oo}{\fn[#1][\mathcal{L}][#2][#3][#4][#5]}
\newcommand*{\steinRKHS}{\setSymbol{H}}
\newcommand*{\lengthscale}{\sigma}

\newcommand*{\dataset}{\mathcal{D}}
\newcommand*{\preds}{\vec{x}}

\newcommand*{\binresp}{y}

\newcommand*{\dataSize}{n}

\newcommand*{\wrt}{w.r.t.\xspace}
\newcommand*{\ie}{i.e.\xspace}
\newcommand*{\eg}{e.g.\xspace}
\newcommand*{\st}{s.t.\xspace}

\newcommand*{\sigopt}{SigSVGD\xspace}

\begin{document}

\maketitle
\thispagestyle{empty}
\pagestyle{empty}

\begin{abstract}
Motion planning can be cast as a trajectory optimisation problem where a cost is minimised as a function of the trajectory being generated.
In complex environments with several obstacles and complicated geometry, this optimisation problem is usually difficult to solve and prone to local minima.
However, recent advancements in computing hardware allow for parallel trajectory optimisation where multiple solutions are obtained simultaneously, each initialised from a different starting point.
Unfortunately, without a strategy preventing two solutions to collapse on each other, naive parallel optimisation can suffer from mode collapse diminishing the efficiency of the approach and the likelihood of finding a global solution.
In this paper we leverage on recent advances in the theory of rough paths to devise an algorithm for parallel trajectory optimisation that promotes diversity over the range of solutions, therefore avoiding mode collapses and achieving better global properties.
Our approach builds on path signatures and Hilbert space representations of trajectories, and connects parallel variational inference for trajectory estimation with diversity promoting kernels.
We empirically demonstrate that this strategy achieves lower average costs than competing alternatives on a range of problems, from 2D navigation to robotic manipulators operating in cluttered environments. 
\end{abstract}

\section{Introduction}

Trajectory optimisation is one of the key tools in robotic motion, used to find control signals or paths in obstacle-cluttered environments that allow the robot to perform desired tasks. 
These trajectories can represent a variety of applications, such as the motion of autonomous vehicles or robotic manipulators. 
In most problems, we consider a \emph{state-space model}, where each distinct situation for the world is called a \emph{state}, and the set of all possible states is called the \emph{state space}~\citep{lavalle_planning_2006}.
When optimising candidate trajectories for planning and control, two criteria are usually considered: \emph{optimality} and \emph{feasibility}.
Although problem dependant, in general, the latter evaluates in a binary fashion whether the paths generated respect the constraints of both the robot and the task, such as physical limits and obstacle avoidance.
Conversely, optimality is a way to measure the quality of the generated trajectories with respect to task-specific desired behaviours.
For example, if we are interested in smooth paths we will search for trajectories that minimise changes in velocity and/or acceleration.
The complexity of most realistic robot planning problems scales exponentially with the dimensionality of the state space and is countably infinite.
When focusing on motion planning, a variety of algorithms have been proposed to find optimal and feasible trajectories.
These can be roughly divided into two main paradigms: sampling-based and trajectory optimisation algorithms.

\begin{figure}[t]
    \centering
    \includegraphics[width=.8\linewidth]{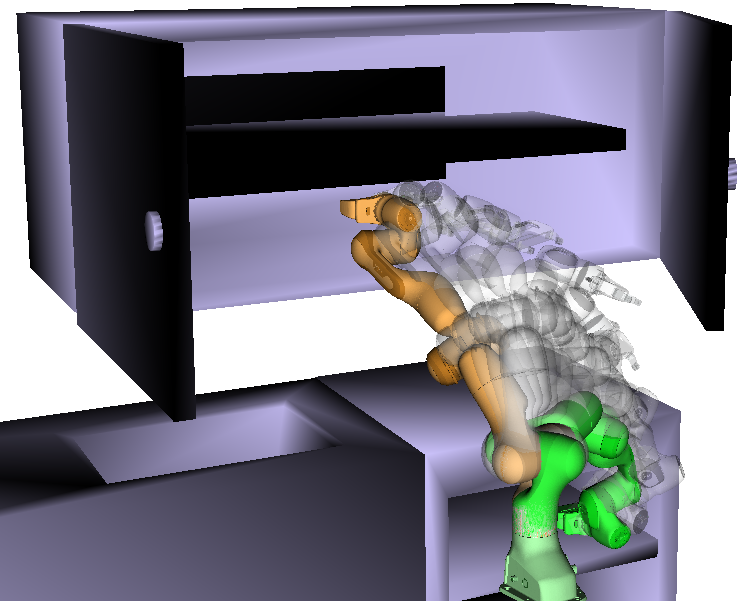}
    \caption{\label{fig:intro-image}
        \textbf{An episode of the \emph{Kitchen} scene.}
        Depicted is one of the collision-free paths found by \sigopt\ on a reaching task using a 7 DOF Franka Panda arm on the MotionBenchMaker planning benchmark.
        }
    \vspace{-10pt}
\end{figure}

Sampling-based planning~\citep{gammell_asymptotically_2021} is a class of planners with \emph{probabilistically complete} and \emph{asymptotically optimal} guarantees~\citep{al-bluwi_motion_2012}.
These approaches decompose the planning problem into a series of sequential decision-making problems with a tree-based~\citep{lavalle_randomized_2001} or graph-based~\citep{kavraki_probabilistic_1996, jaillet_path_2008} approach.
However, most approaches are limited in their ability to encode kinodynamic cost like trajectory curvature~\citep{heilmeier_minimum_2020} or acceleration torque limits~\citep{berntorp_models_2014}.
In addition, despite the completeness guarantee, sampling-based planners are often more computationally expensive as the search space grows and can obtain highly varying results due to the random nature of the algorithms.

Trajectory optimisation algorithms~\citep{gonzalez_review_2016} use different techniques to minimise a cost functional that encourages solutions to be both optimal and feasible.
The most direct optimisation procedure relies on a differentiable cost function and uses functional gradient techniques to iteratively improve the trajectory quality~\citep{ratliff_chomp_2009}.
However, many different strategies have been proposed.
For example, one may start from a randomly initialised candidate trajectory and proceed by adding random perturbations to explore the search space and generate approximate gradients, allowing any arbitrary form of cost functional to be encoded~\citep{kalakrishnan_stomp:_2011}.
The same approach can be used to search for control signals and a local motion plan concurrently~\citep{williams_aggressive_2016}.
Finally, a locally optimal trajectory can also be obtained via decomposing the planning problem with sequential quadratic programming~\citep{schulman_finding_2013}.
A drawback of these methods is that they usually find solutions that are locally optimal and may need to be run with different initial conditions to find solutions that are feasible or with lower costs.

\begin{figure*}
    \centering
    \includegraphics[width=\linewidth]{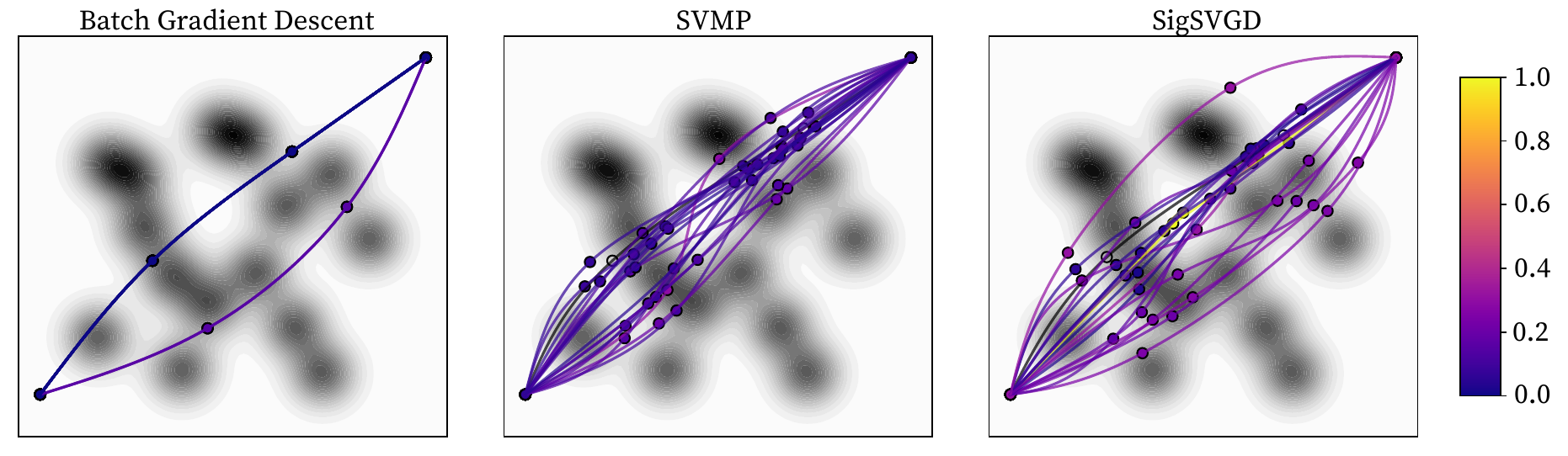}
    \caption{
        \textbf{Qualitative analysis of 2D planning task.}
        The plot shows the final 20 trajectories found with different optimisation methods.
        The colour of each path shows its normalised final cost.
        Note how all batch gradient descent trajectories converge to two modes of similar cost.
        Paths found by SVMP are already more diverse, but one of the gradient descent modes is lost.
        Note how when multiple trajectories converge to a single trough, the knots are pushed away by the repulsive force resulting in suboptimal solutions.  
        Conversely, paths found by \sigopt\ are diverse and able to find more homotopic solutions, including those found by BGD.
        Note also how paths are able to converge to the same trough without being repelled by one another since the repulsive force takes into account the entire trajectory and not exclusively the spline knot placement.
        That also allows for paths that are more direct and coordinated than SVMP.
    }\label{fig:2d_planning}
\end{figure*}

Our goal with the present work is to propose a new trajectory optimisation method to improve path diversity.
More specifically, we focus on a class of algorithms that perform trajectory optimisation parallel optimisation of a batch of trajectories.
This concurrent optimisation of several paths in itself already alleviates the proneness to local minima, since many initial conditions are evaluated simultaneously.
Nonetheless, we show how a proper representation of trajectories when performing functional optimisation leads to increased diversity and solutions with a better global property, either with direct gradients or Monte Carlo-based gradient approximations.
As an illustrative example, refer to~\cref{fig:2d_planning}.

Our approach is based on two cornerstones.
On one hand, we use a modification of Stein Variational Gradient Descent (SVGD)~\citep{liu_stein_2016}, a variational inference method to approximate a posterior distribution with an empirical distribution of sampled particles, to optimise trajectories directly on a structured Reproducing Kernel Hilbert Space (RKHS).

The structure of this space is provided by the second pillar of our approach.
We leverage recent advancements in rough path theory to encode the sequential nature of paths in the RKHS using a Path Signature Kernel~\citep{kiraly_kernels_2019,salvi_signature_2021}.
Therefore we can approximate the posterior distribution over optimal trajectories with structured particles during the optimisation while still taking into account motion planning and control idiosyncrasies.

More concretely, the main contributions of this paper are listed below:
\begin{itemize}
    \item We introduce the use of path signatures~\citep{lyons_rough_2014} as a canonical feature map to represent trajectories over high-dimensional state spaces;
    \item Next, we outline a procedure to incorporate the signature kernel into a variational inference framework for motion planning;
    \item Finally, we demonstrate through experiments in both planning and control that the proposed procedure results in more diverse trajectories, which aid in avoiding local minima and lead to a better optimisation outcome. 
\end{itemize}

The paper is organised as follows. In~\Cref{sec:related} we review
related work, contrasting the proposed method to the existing literature.
In~\Cref{sec:background} we provide background on path signatures and motion planning as variational inference, which are the foundational knowledge for the method outlined in~\Cref{sec:method}.
Finally, in~\Cref{sec:results} we present a number of simulated experiments, followed by relevant discussions in~\Cref{sec:conclusion}.

\section{Related Work}\label{sec:related}

Trajectory optimisation refers to a class of algorithms that start from an initial sub-optimal path and find a, possibly local, optimal solution by minimising a cost function.
Given its broad definition, there are many seminal works in the area. 
One influential early work is Covariant Hamiltonian Optimisation for Motion Planning (CHOMP)~\citep{ratliff_chomp_2009} and related methods~\citep{zucker_chomp_2013,byravan_space-time_2014,marinho_functional_2016}. The algorithm leverages the covariance of trajectories coupled with Hamiltonian Monte Carlo to perform annealed functional gradient descent.
However, one of the limitations of CHOMP and related approaches is the need for a fully-differentiable cost function.

In Stochastic Trajectory Optimisation for Motion Planning (STOMP)~\citep{kalakrishnan_stomp:_2011} the authors address this by approximating the gradient from stochastic samples of noisy trajectories, allowing for non-differentiable costs.
Another approach used in motion planning are quality diversity algorithms, at the intersection of optimisation and evolutionary strategies, of which Covariance Matrix Adaptation Evolution Strategy (CMA-ES) is the most prominent~\citep{hansen_reducing_2003,hamalainen_ppo-cma_2020,tjanaka_training_2022}.
CMA-ES is a derivative-free method that uses a multivariate normal distribution to generate and update a set of candidate solutions, called individuals.
The algorithm adapts the covariance matrix of the distribution based on the observed fitness values of the individuals and the search history, balancing exploration and exploitation of the search space. 
Because of its stochastic nature, it is ergodic and copes well with multi-modal problems.
Nonetheless, it may require multiple initialisations and it typically requires more evaluations than gradient-based optimisers~\citep{hansen_cma_2016}.

TrajOpt \citep{schulman_finding_2013}, another prominent planner, adopts a different approach solving a sequential quadratic program and performing continuous-time collision checking.
Contrary to sampling-based planners, these trajectory optimisation methods are fast, but only find locally optimal solutions and may require reiterations until a feasible solution is found.
Another issue common to these approaches is that in practice they require a fixed and fine parametrisation of trajectory waypoints to ensure feasibility and smoothness, which negates the benefit of working on continuous trajectory space.
To address this constraint, in~\citep{marinho_functional_2016} the authors restrict the optimisation and trajectory projection to an RKHS with an associated squared-exponential kernel.
However, the cost between sparse waypoints is ignored and the search is still restricted to a deterministic trajectory.
Another approach was proposed in GPMP~\citep{dong_motion_2016,mukadam_simultaneous_2017,mukadam_continuous-time_2018} by representing trajectories as Gaussian Processes (GP) and looking for a \emph{maximum a posteriori} (MAP) solution of the inference problem.

More closely related to our approach are~\citep{lambert_entropy_2021,yu_gaussian_2022} which frame motion planning as a variational inference problem and try to estimate the posterior distribution represented as a set of trajectories.
In~\citep{yu_gaussian_2022}, the authors modify GPMP with a natural gradient update rule to approximate the posterior.
On the other hand, in Stein Variational Motion Planning (SVMP)~\citep{lambert_entropy_2021} the posterior inference is optimised using Stein variational gradient descent.
This method is similar to ours, but the induced RKHS does not take into account the sequential nature of the paths being represented, which leads to a diminished repulsive force and lack of coordination along the dimensions of the projected space.

In contrast, our approach---which we will refer to as Kernel Signature Variational Gradient Descent (\sigopt)---uses the path signature to encode the sequential nature of the functional being optimised.
We argue that this approach leads to a better representation of trajectories promoting diversity and finding better local solutions.
To empirically corroborate this claim we use the Occam's razor principle and take SVMP as the main baseline of comparison since it more closely approximates our method.

We note that the application of trajectory optimisation need not be restricted to motion planning.
By removing the constraint of a target state and making the optimisation process iterative over a rolling horizon we retrieve a wide class of Model Predictive Controllers with applications in robotics~\citep{williams_aggressive_2016,barcelos_disco_2020,barcelos_dual_2021,lambert_stein_2020}.
Stein Variational MPC (SVMPC)~\citep{lambert_stein_2020} uses variational inference with SVGD optimisation to approximate a posterior over control policies and more closely resembles \sigopt.
However, like SVMP, it too does not take into account the sequential nature of control trajectories and we will illustrate how our approach can improve the sampling of the control space and promote better policies. 

\section{Background}\label{sec:background}

\begin{figure*}
    \centering
    \includegraphics{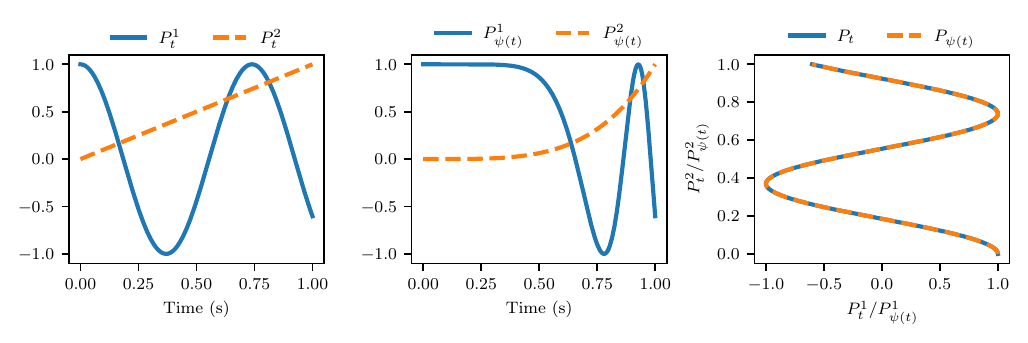}
    \caption{\label{fig:pathsig_invariance}
        \textbf{Signature invariance to reparametrisation.}
        \emph{Left}: Plot of the coordinates of a two dimensional path \( P_{t} \) over time. Here \( P_{t}^{1} = \cos (8.5 t)  \) and \( P_{t}^{2} = t \).
        \emph{Centre}: Plot of the two coordinates of path \( P_{t} \) reparameterised by function \( \psi \). Now, \( P_{\psi(t)}^{1} = \cos (8.5 t^{4})  \) and \( P_{\psi(t)}^{2} = t^{4} \). 
        \emph{Right}: Plots of path \( P_{t} \) and its reparameterised version \( P_{\psi(t)} \) are shown overlapping to illustrate how the change in time is irrelevant if the goal is achieving diverse paths. The signature of degree 2 for both paths is \( \{1, -1.6,  1,  1.3, -0.9, -0.7,  0.5\} \).
    }
\end{figure*}

\subsection{Trajectory Optimisation in Robotics}\label{sec:prelim}
Consider a system with state $\state \in \stateSpace$ and let us denote a \emph{trajectory} of such system as $\stateSeq: [\llim, \ulim] \to \stateSpace$, where $\stateSpace$ is an appropriate Euclidean space or group. We shall use the notation $\stateSeq_{t}$ to denote the dependency on time $t \in [\llim, \ulim]$.
The trajectory \( \stateSeq \) describes a \emph{path} in \( \stateSpace \) and we shall use the two denominations interchangeably.
In trajectory optimisation the goal is to find the optimal path \( \stateSeq^{*} \) from a given starting state $\stateInit$ to a certain goal state $\stateGoal$. This can be done by minimising a cost functional that codifies our desired behaviour $\costFn \colon \trajSpace \to \R^{+}$, where $\trajSpace$ is the Hilbert space of trajectories~\citep{king_pregrasp_2013}:
\begin{equation}\label{eq:traj_optim_prob}
    \stateSeq^{*} \coloneqq \argmin_{\stateSeq} \costFn[\stateSeq],
    \text{~\st~} \stateSeq_\llim = \stateInit \; \textup{and} \; \stateSeq_\ulim = \stateGoal.
\end{equation}
Typically, \( \costFn \) is a bespoke functional that includes penalties for trajectory non-smoothness, total energy, speed and acceleration tracking, as well as length. To ensure that the solution is feasible and collision-free, additional equality and inequality constraints may also be included~\citep{schulman_finding_2013}. 
Alternatively, we can solve an unconstrained problem and include additional penalties to the cost functional as soft-constraints~\citep{zucker_chomp_2013, ratliff_chomp_2009}.

Finally, we draw the reader's attention to the fact that the problem stated in~\cref{eq:traj_optim_prob} can be viewed as an open-loop optimal control problem. If the solution can be found in a timely manner, the same problem can be cast onto a Model Predictive Control~\citep{camacho_model_2013, barcelos_disco_2020, barcelos_dual_2021} framework
\begin{equation}\label{eq:control_prob}
    \ctrlSeq^{*} \coloneqq \argmin_{\ctrlSeq} \costFn[\stateSeq, \ctrlSeq],
    \text{~\st~} \stateSeq_\llim = \stateInit,
\end{equation}
where \( \ctrlSeq: [\llim, \ulim] \to \ctrlSpace \) is a path of control inputs on a given Euclidean space and the mapping to \( \stateSpace \) is given by the dynamical system \( \transFn \) such that \( \dot{\state} = \transFn (\state, \ctrl, t) \). That is to say, we now influence the path \( \stateSeq \) indirectly through input \( \ctrlSeq \), and at any time \( t \) the problem is solved for a finite interval. The closed-loop solution arises from applying only the first immediate control action before re-optimising the solution.

\subsection{Path Signature}\label{sec:pathsig}
A multitude of practical data streams and time series can be regarded as a path, for example, video, sound, financial data, control signals, handwriting, etc. The path signature transforms such multivariate sequential data (which may have missing or irregularly sampled values) into an infinite-length series of real numbers that uniquely represents a trajectory through Euclidean space. Although formally distinct and with notably different properties, one useful intuition is to think of the signature of a path as akin to a Fourier transform, where paths are summarised by an infinite series of feature space coefficients.
Consider a path $\anypath$ traversing space $\stateSpace \subseteq \R^{\sigChannels}$ as defined in~\cref{sec:prelim}. Note that at any time \( t \) such path can be decomposed in $\anypath_{t} = \left\{ \anypath_{t}^{1}, \anypath_{t}^{2}, \ldots, \anypath_{t}^{\sigChannels} \right\}$.
Now recall that for a one-dimensional path $\anypath_{t}$ and a function $\anyfunction$, the path integral of $\anyfunction$ along $\anypath$ is defined by:
\begin{equation}\label{eq:path_integral}
    \int_{\llim}^{\ulim} \anyfunction (\anypath_{t}) \dif \anypath_{t} = \int_{\llim}^{\ulim} \anyfunction (\anypath_{t}) \dot{\anypath_{t}} \dif t .
\end{equation}

In particular, note that the mapping $t \to \anyfunction (\anypath_t)$ is also a path.
In fact, \cref{eq:path_integral} is an instance of the Riemann-Stieltjes integral~\citep{chevyrev_primer_2016}, which computes the integral of one path against another.
Let us now define the\emph{ 1-fold iterated} integral, which computes the increment of the $\primIdx$-th coordinate of the path at time \(t\) as:
\begin{equation}
    {\signature*[\anypath]}_{t}^{\primIdx} =
    \int\mclimits_{\llim < t_1 < t} \dif \anypath_{t_1}^{\primIdx} =
    \anypath_{t}^{\primIdx} - \anypath_{\llim}^{\primIdx} ,
\end{equation}
and we again emphasise that \( {\signature[\anypath]}_{t}^{\primIdx} \) is also a real valued path. This allows us to apply the same iterated integral recursively and we proceed by defining the \emph{2-fold iterated} integral~\citep{chen_iterated_1954, chen_iterated_1977} as:
\begin{equation}
    {\signature*[\anypath]}_{t}^{\primIdx, \secIdx} = \int\mclimits_{\llim < t_2 < t}
    {\signature*[\anypath]}_{ t_2}^{\primIdx} \dif \anypath_{t_2}^{\secIdx} =
    \int\mclimits_{\llim < t_1 < t_2 < t} \dif \anypath_{t_1}^{\primIdx} \dif \anypath_{t_2}^{\secIdx} .
\end{equation}
Informally, we can proceed indefinitely and we retrieve the path signature by collecting all iterated integrals of the path \( \anypath \). A geometric intuition of the signature can be found in~\citep{chevyrev_primer_2016, yang_developing_2017} where the first three iterated integrals represent displacement, the L\'{e}vy area~\citep{lyons_differential_2007} and volume of the path respectively.

\begin{definition}[Signature~\citep{chevyrev_primer_2016}]
    The \emph{signature} of a path \( \anypath: t \in [\llim, \ulim] \to \R^{\sigChannels} \), denoted by \( {\signature[\anypath]}_{t} \), is the infinite series of all iterated integrals of \( \anypath \). Formally, \( {\signature[\anypath]}_{t} \) is the sequence of real numbers
    \begin{equation}
        {\signature*[\anypath]}_{t} = \bigl(
            1, {\signature*[\anypath]}_{t}^{1}, \ldots, {\signature*[\anypath]}_{t}^{\sigChannels},
            {\signature*[\anypath]}_{t}^{1, 1}, {\signature*[\anypath]}_{t}^{1, 2}, \ldots  
        \bigr) ,
    \end{equation}
    where the iterated integrals are defined as:
    \begin{equation}
        {\signature*[\anypath]}_{t}^{\primIdx_{1}, \ldots, \primIdx_{\thirdIdx}} = \int\limits_{\llim < t_{\thirdIdx} < t}
        \ldots \int\limits_{{\llim < t_{1} < t_{2}}} \dif \anypath_{t_1}^{\primIdx_{1}} \ldots 
        \dif \anypath_{t_\thirdIdx}^{\primIdx_\thirdIdx} ,
    \end{equation}
    and the superscripts are drawn from the set \( \setSymbol{M} \) of all multi-indexes,
    \begin{equation}
        \setSymbol{M} = \bigl\{ (\primIdx_1, \ldots, \primIdx_\thirdIdx) \given \thirdIdx \geq 1, \primIdx_1, \ldots, \primIdx_\thirdIdx \in \{ 1, \ldots, \sigChannels \}   \bigr\}.
    \end{equation}
\end{definition}

In practice we often apply a truncated signature up to a degree \( \sigDegree \), that is \( {\signature^{\sigDegree}[\anypath]}_{t} \), defined as the finite collection of all terms of the signature up to multi-indices of length \( \sigDegree \).

The path signature was originally introduced by Chen~\citep{chen_integration_1958} who applied it to piecewise smooth paths and further developed by Lyons and others~\citep{amendola_varieties_2019, boedihardjo_signature_2016, hambly_uniqueness_2010, lyons_rough_2014}.
The number of elements in the path signature depends on the dimension of the input \( \sigChannels \) and the degree \( \sigDegree \), and is given by \( \sigChannels^{\sigDegree} \).
Therefore the time and space scalability of the signature is rather poor (\( O(\sigChannels^\sigDegree) \)), but this can be alleviated with the use of kernel methods as we will discuss in~\cref{sec:method}.
The signature of a path has several interesting properties which make it inherently interesting for applications in robotics.

\subsubsection*{Canonical feature map for paths:} For all effects, the path signature can be thought of as a \emph{linear} feature map~\citep{fermanian_embedding_2021} that transforms multivariate sequential data into an infinite length series of real numbers which uniquely represents a trajectory through Euclidean space.
This is valid even for paths with missing or irregularly sampled values~\citep{boedihardjo_signature_2016, hambly_uniqueness_2010}.

\subsubsection*{Time-reversal:} We informally define the time-reversed path \( \overleftarrow{\anypath} \) as the original path \( \anypath \) moving backwards in time. It follows that the tensor product of the signatures \( {\signature[\anypath]}_{\llim,\ulim} \otimes {\signature[\overleftarrow{\anypath}]}_{\llim,\ulim} = 1 \), which is the identity operation.

\subsubsection*{Uniqueness:} The signature of every non tree-like path is unique~\citep{hambly_uniqueness_2010}. A tree-like path is one in which a section exactly retraces itself. Tree-like paths are quite common in real data (\eg\ in cyclic actions) and this could be a limiting factor of the signature's application. However, it has been proven~\citep{hambly_uniqueness_2010} that if a path has at least one monotonous coordinate, then its signature is unique. The main significance of this result is that it provides a practical procedure to guarantee signature uniqueness by, for example, including a time dimension.

\subsubsection*{Invariance under reparametrisation:} An important difficulty when vying for diversity in trajectory optimisation is the potential symmetry present in the data. This is particularly true when dealing with sequential data, such as, for instance, trajectories of an autonomous vehicle. In this case, the problem is compounded as there is an infinite group of symmetries given by the reparametrisation of a path (\ie\ continuous surjections in the time domain to itself), each leading to distinct similarity metrics. In contrast, the path signature acts as a filter that is invariant to reparametrisation removing these troublesome symmetries and resulting in the same features as shown in~\Cref{fig:pathsig_invariance}.

\subsubsection*{Dimension is independent of path length:} The final property we will emphasise is how the dimension of the signature depends on its degree and the intrinsic dimension of the path, but is independent of the path length. In other words, the signature dimension is invariant to the degree of discretisation of the path.

\subsection{Stein Variational Gradient Descent}\label{sec:svgd}

Variational inference (VI)~\citep{blei_variational_2017} is an established and powerful method for approximating challenging posterior distributions in Bayesian Statistics. As opposed to Markov chain Monte Carlo (MCMC)~\citep{haugh_tutorial_2021} approaches, in VI the inference problem is cast as an optimisation problem in which a candidate distribution $\qPdf^{*}[\anyvector]$ within a distribution family $\qClass$ is chosen to best approximate the target distribution $\pPdf[\anyvector]$.  This is typically obtained by minimising the Kullback-Leibler (KL) divergence:
\begin{equation} \label{eq:vi_obj}
    \qPdf^{*} = \argmin_{\qPdf \in \qClass} \
    \kl {\qPdf} {\pPdf} .
\end{equation}
The solution also maximises the Evidence Lower Bound (ELBO), as expressed by the following objective
\begin{equation}
    \qPdf^{*} = \argmax_{\qPdf \in \qClass}
    \expectation_{\qPdf}[\big][\log \pPdf[\anyvector]]
    - \kl {\qPdf[\anyvector]} {\pPdf[\anyvector]} .
\end{equation}

The main challenge that arises is defining an appropriate $\qClass$. Stein variational gradient descent (SVGD)~\citep{liu_stein_2016} addresses this issue while also solving for \cref{eq:vi_obj} by performing Bayesian inference in a non-parametric nature, removing the need for assumptions on restricted parametric families for $\qPdf [\anyvector]$. This approach approximates a posterior $\pPdf [\anyvector]$ with a set of particles ${\{\anyvector^\primIdx\}}_{\primIdx = 1}^{\partSize}$, $\anyvector \in \R^{\partDim}$. These particles are iteratively updated in parallel according to: 
\begin{equation}\label{eq:stein_update}
    \anyvector^{\primIdx} \leftarrow \anyvector^{\primIdx} + 
    \stepSize \scoreFunc^{*}[\anyvector^{\primIdx}],
\end{equation} 
given a step size $\stepSize$.
The function $\scoreFunc[\cdot]$ is known as the score function and defines the velocity field that maximally decreases the KL-divergence:
\begin{equation}\label{eq:stein_optim}
    \scoreFunc^{*} = 
    \argmax_{\scoreFunc \in \steinRKHS}~\bigl\{
        -\grad_{\stepSize} \kl{\qPdf_{[\stepSize \scoreFunc]}}{\pPdf},
        \text{~\st~} \|\scoreFunc\|_{\steinRKHS} \leq 1
    \bigr\},
\end{equation}
where $\steinRKHS$ is a Reproducing Kernel Hilbert Space (RKHS) induced by a positive-definite kernel $k:\stateSpace\times\stateSpace\to\R$, and $\qPdf_{[\stepSize \scoreFunc]}$ indicates the particle distribution resulting from taking an update step as in \cref{eq:stein_update}.
Recall that an RKHS $\steinRKHS$ associated with a kernel $k$ is a Hilbert space of functions endowed with an inner product $\inner{\cdot}{\cdot}$ such that $f(\anyvector) = \inner{f}{k(\cdot, \anyvector)}$ for any $f \in \steinRKHS$ and any $\anyvector\in\stateSpace$ \citep{scholkopf_learning_2002}.
In~\citep{liu_stein_2016}, the problem in \eqref{eq:stein_optim} has been shown to yield a closed-form solution which can be interpreted as a functional gradient in $\steinRKHS$ and approximated with the set of particles:
\begin{equation} \label{eq:stein_score_func}
    \scoreFunc^{*} [\anyvector] = \expectation_{\othervector \sim \hat{\qPdf}} [\big]
    [k \parens{\othervector, \anyvector} \grad_{\othervector} \log \pPdf[\othervector] + \grad_{\othervector} k \parens{\othervector, \anyvector}],
\end{equation}
with \( \hat{\qPdf} = \frac{1}{\partSize} \sum_{\primIdx = 1}^{\partSize} \dirac{\anyvector^{\primIdx}} \) being an empirical distribution that approximates \( \qPdf \) with a set of Dirac delta functions \( \dirac{\anyvector^{\primIdx}} \). For SVGD, $k$ is typically a translation-invariant kernel, such as the squared-exponential or the Mat\'ern kernels \citep{liu_stein_2016, rasmussen_gaussian_2006}.

\section{Method}\label{sec:method}
Our main goal is to find a diverse set of solutions to the problem presented in~\cref{sec:prelim}.
To that end, we begin by reformulating~\cref{eq:traj_optim_prob} as a probabilistic inference problem.
Next, we show that we can apply SVGD to approximate the posterior distribution of trajectories with a set of sampled paths.
Finally, in~\cref{sec:sig_svgd}, we present our main contribution discussing how we can promote diversity among the sample paths by leveraging the Path Signature Kernel.

\subsection{Stein Variational Motion Planning}\label{sec:svgd_mp}
To reframe the trajectory optimisation problem described in~\cref{eq:traj_optim_prob} as probabilistic inference we introduce a binary optimality criterion, \( \optimality: \trajSpace \to \{0, 1\} \), analogously to~\citep{barcelos_dual_2021, levine_reinforcement_2018}.
Simplifying the notation with \( \optimality \) indicating \( \optimality~=~1 \), we can represent the posterior distribution of optimal trajectories as \( \pPdf[\anypath \given \optimality] \propto \pPdf[\optimality \given \anypath] \pPdf[\anypath] \), for a given optimality likelihood \( \pPdf[\optimality \given \anypath] \) and trajectory prior \( \pPdf[\anypath] \).
The \emph{maximum a posteriori} (MAP) solution is given by finding the mode of the negative log posterior:
\begin{equation}
    \label{eq:mp_as_pi}
    \begin{aligned}
        \anypath^{*} &= \argmin_{\anypath} -\log  \pPdf[\optimality \given \anypath] - \log \pPdf[\anypath] \\
                     &= \argmin_{\anypath} \temperature \costFn[\anypath] - \log \pPdf[\anypath],
    \end{aligned}
\end{equation}
where the last equality arises from the typical choice of the exponential distribution to represent the optimality likelihood, \ie\ \( \pPdf[\optimality \given \anypath] = \exp (-\temperature \costFn[\anypath]) \) with \( \temperature \) being a temperature hyper-parameter.

Rather than finding the MAP solution, we are interested in approximating the full posterior distribution, which may be multi-modal, and generating diverse solutions for the planning problem.
As discussed in~\cref{sec:svgd}, we can apply SVGD to approximate the posterior distribution with a collection of particles.
In the case at hand each of such particles is a sampled path, such that \cref{eq:stein_score_func} can be rewritten as:
\begin{equation}\label{eq:svmp_score_func}
    {\scoreFunc}^{*} (\anypath) = \expectation_{\otherpath \sim \hat{\qPdf}} [\big] 
    [k \parens{\otherpath, \anypath} \grad_{\otherpath} \log \pPdf[\otherpath \given \optimality] + \grad_{\otherpath} k \parens{\otherpath, \anypath}].
\end{equation}

The score function presented in~\cref{eq:svmp_score_func} is composed of two competing forces.
On one hand, we have the kernel smoothed gradient of the log-posterior pushing particles towards regions of higher probability.
Whereas the second term acts as a repulsive force, pushing particles away from one another.

It is worth emphasising that the kernel function is \emph{static}, \ie\ it does not consider the sequential nature of the input paths.
In effect, for a path of dimension \( \sigChannels \) and \( \anyscalar \) discrete time steps, the inputs are projected onto a space \( \setSymbol{V} \subset \R^{\sigChannels \times \anyscalar} \) in which similarities are evaluated.

Finally, the posterior gradient can be computed by applying Bayes' rule, resulting in:
\begin{equation}\label{eq:svmp_grad}
    \grad_{\otherpath} \log \pPdf[\otherpath \given \optimality] =  \grad_{\anypath} \log \pPdf[\otherpath] - \grad_{\otherpath} \temperature \costFn[\otherpath].
\end{equation}

\subsection{Stein Variational Motion Planning with Smooth Paths}\label{sec:svmp_on_splines}
In previous work~\citep{barfoot_batch_2014, dong_motion_2016, lambert_entropy_2021, mukadam_continuous-time_2018} the prior distribution in~\cref{eq:svmp_grad} is defined in a way to promote smoothness on generated paths.
This typically revolves around defining Gaussian Processes~\citep{rasmussen_gaussian_2006} as priors and leveraging factor graphs for efficiency.
Although effective, this approach still requires several latent variables to describe a desired trajectory, which implies on a higher dimensional inference problem.

Importantly, the problem dimensionality is directly related to the amount of repulsive force exerted by the kernel.
In large dimensional problems, the repulsive force of translation-invariant kernels vanishes, allowing particles to concentrate around the posterior modes which results in an underestimation of the posterior variance~\citep{zhuo_message_2018}.
This problem is further accentuated when considering the static nature of the kernel function, as discussed in the previous section.

In order to keep the inference problem low-dimensional while still enforcing smooth paths we make use of \emph{natural cubic splines} and aim to optimise the location of a small number of knots.
These knots may be initialised in different ways, such as perturbations around a linear interpolation from the starting state \( \stateInit \) and goal state \( \stateGoal \), sampled from an initial solution given by a shooting method (\eg\ RRT~\citep{lavalle_randomized_2001}), or drawn randomly from within the limits of \( \stateSpace \).
For simplicity, in this work we will opt for the latter.

Since path smoothness is induced by the splines, the choice of prior is more functionally related to the problem at hand.
If one desires some degree of regularisation on the trajectory optimisation, a multivariate Gaussian prior centred at the placement of the initial knots may be used.
Conversely, if we only wish to ensure the knots are within certain bounds, a less informative smoothed approximation of the uniform prior may be used. More concretely, for a box \( B = {x\colon \llim \leq x \leq \ulim}\), such prior would be defined as:
\begin{equation}
    \pPdf[x] \propto \exp{\left(- d{\left(x, B\right)}^2 / \sqrt{(2 \sigma^2)}\right)}
\end{equation}
where the distance function \( d\left(x, B\right) \) is given by $ d\left(x, B\right) = \min |x - x'|, \; x' \in B $.
Finally, we could define both a prior and hyper-prior if we wish to combine both effects (see~\cref{app:hyperprior} for details).

As discussed in~\cref{sec:prelim} the cost functional \( \costFn \) imposes penalties for collisions and defines the relevant performance criteria to be observed.
Since only a small number of knots is used for each path, some of these criteria and, in particular, collision checking require that we discretise the resulting spline in a sufficiently dense amount of points.
It is worth mentioning that \( \costFn \) is typically non-differentiable and that the gradient in~\cref{eq:svmp_grad} is usually approximated with Monte Carlo samples~\citep{barcelos_dual_2021}.
However, as this introduces an extra degree of stochasticity in the benchmark comparison, we will restrict our choice of \( \costFn \) to be differentiable.
We will discuss the performance criteria of each problem in the experimental section.

\SetKwComment{Comment}{-- }{}
\begin{algorithm*}
    \caption{Kernel Signature Stein Variational Gradient Descent (\sigopt)}
    \label{algo:sigsvgd}
    \KwIn{
        A cost function $\costFn[\stateSeq]$ or target distribution $\pPdf[\stateSeq]$,
        a prior distribution $\qPdf[\stateSeq_{t_0}]$,
        a signature kernel $k^{\oplus}$.
    }
    \KwOut{
        A set of particles $\set*{\stateSeq_{t}^{\primIdx}}_{\primIdx = 1}^{\partSize}$ that approximates the posterior distribution over optimal paths.
    }

    Sample $\set*{\stateSeq_{t_0}^{\primIdx}}_{\primIdx = 1}^{\partSize} \sim \qPdf[\stateSeq_{t_0}]$\;

    %
    \While{task not complete}{
        \If{using Monte Carlo samples}{
            Generate $\nSamples$ samples for each path $\stateSeq_{t}^{\primIdx, \secIdx} \gets \stateSeq_{t}^{\primIdx} + \noise_{\secIdx}$\;
        }
        \If{using splines}{
            Generate decimated trajectories from knots $\stateSeq_{t}$\;
        }
        Evaluate $\costFn[\stateSeq_{t}]$ in parallel\;
        \If{target distribution \( \pPdf[\stateSeq_{t}] \) is available}{
            Update score $\scoreFunc^{*} \gets \frac{1}{\partSize} \sum_{\primIdx}^{\partSize} \bracks{k^{\oplus}
                \parens{\stateSeq_{t}^{\primIdx}, \stateSeq_{t}} \grad_{\stateSeq_{t}^{\primIdx}} \log \pPdf[\stateSeq_{t}^{\primIdx}]
            + \grad_{\stateSeq_{t}^{\primIdx}} k^{\oplus} \parens{\stateSeq_{t}^{\primIdx}, \stateSeq_{t}}}$\;
        }
        \Else{
            Log-posterior gradient $\grad_{\stateSeq_{t}^{\primIdx}} \log \pPdf[\stateSeq_{t}^{\primIdx} \given \optimality]
            \approx \grad_{\stateSeq_{t}^{\primIdx}} \log \qPdf[\stateSeq_{t-1}^{\primIdx} \given \optimality]
            + \grad_{\stateSeq_{t}^{\primIdx}} \log \frac{1}{\nSamples} \sum_{\secIdx}^{\nSamples} \exp(-\alpha \costFn[\stateSeq_{t}^{\primIdx, \secIdx}])$\;
            Update score $\scoreFunc^{*} \gets \frac{1}{\partSize} \sum_{\primIdx}^{\partSize} \bracks{k^{\oplus}
                \parens{\stateSeq_{t}^{\primIdx}, \stateSeq_{t}} \grad_{\stateSeq_{t}^{\primIdx}} \log \pPdf[\stateSeq_{t}^{\primIdx} \given \optimality]
            + \grad_{\stateSeq_{t}^{\primIdx}} k^{\oplus} \parens{\stateSeq_{t}^{\primIdx}, \stateSeq_{t}}}$\;
        }
        Update paths $\stateSeq_{t} \gets \stateSeq_{t} + \stepSize \scoreFunc^{*}$\;
        Update prior $\qPdf[\stateSeq_{t} \given \optimality] \gets \pPdf[\stateSeq_{t} \given \optimality]$ \Comment*{For details, see~\citep{barcelos_dual_2021}}
        $t \gets t + 1$\;
    }
\end{algorithm*}

\subsection{Stein Variational Motion Planning with Path Signature Kernel}\label{sec:sig_svgd}
In this section we present our main contribution, which is a new formulation for motion planning in which Path Signature can be used to efficiently promote diversity in trajectory optimisation through the use of Signature Kernels. 
In \cref{sec:background} we discussed some desirable properties of the signature transform. 
The key insight is that the space of linear combination of signatures forms an algebra, which enables it as a faithful feature map for trajectories~\citep{kiraly_kernels_2019}.

With that in mind, perhaps the most straightforward use of the signature would be to redefine the kernel used in~\cref{eq:stein_optim,eq:stein_score_func} as \( \bar{k} \parens*{\anypath, \otherpath} = k \parens*{{\signature[\anypath]}_{t}, {\signature[\otherpath]}_{t}} \).
However, as seen in~\cref{sec:background},  this approach would not be scalable given the exponential time and space complexity of the signature \wrt\ to its degree.
A single evaluation of the Gram kernel matrix for \( \bar{k} \) would be an operation of order \( O(\sigBatch^2 \cdot \sigChannels^{\sigDegree}) \), where \( \sigBatch \) is the number of concurrent paths being optimised, \( \sigDegree \) is the degree of the signature, and \( \sigChannels \) is the dimensionality of the space \( \trajSpace \ni \anypath, \otherpath \).
Furthermore, kernel \( \bar{k} \) is static in the sense that it does not take into account the sequential nature of its domain.
Rather than a kernel \( k \colon \stateSpace \times \stateSpace \to \R \), we want to define a kernel \( k^{+} \colon \trajSpace \times \trajSpace \to \R \), which takes into account the structure induced by paths.

Hence, we take a different approach and proceed by first projecting paths to an RKHS onto which we will then compute the signature.
That is, given a kernel \( k^{+} \colon \trajSpace \times \trajSpace \to \R \), a path \( \anypath \in \trajSpace \) can be lifted to a path in  the RKHS \( \setSymbol{P}_{\steinRKHS} \) through the map \( k_{\anypath}\colon t \mapsto k \parens{\anypath_{t}, \cdot} \), where \( \setSymbol{P}_{\steinRKHS} \) is the set of \( \steinRKHS \)-valued paths.
Finally, we compute the signature of the lifted path \( \signature \parens*{k_{\anypath}}_{t} \) and use it as our final feature map.

At first glance, this further deteriorates scalability, since most useful \( \setSymbol{P}_\steinRKHS \) are infinite dimensional, rendering this approach infeasible.
However, results presented by~\citet[Corollary 4.9]{kiraly_kernels_2019} show that this approach can be completely kernelised.
This allows them to define a \emph{truncated signature kernel}, ${k}^{+}\colon \parens{\anypath_{t}, \otherpath_{t}} \mapsto \inner{\signature^{\sigDegree} \parens{k_{\anypath}}_{t}}{\signature^{\sigDegree} \parens{k_{\otherpath}}_{t}}$, that can be efficiently computed using only evaluations of a static kernel \( k \parens{\anyvector, \othervector} \) at discretised timestamps.
The number of evaluations depends on the truncation degree \( \sigDegree \) and number of discretised steps \( \sigLength \).
Several algorithmic approaches are considered in~\citep{kiraly_kernels_2019} with dynamic programming having complexity \( O \parens{\sigBatch^2 \cdot \sigLength^2 \cdot \sigDegree} \) to compute a \( \parens{\sigBatch \times \sigBatch} \)-Gram matrix.
Otherwise, approximations can be used to reduce the complexity to linear on \( \sigLength \) and \( \sigBatch \).
However, even though the importance of the terms in the signature decay factorially~\citep{lyons_rough_2014}, the amount of coefficients grows exponentially, which means that for high values of \( \sigDegree \) the kernel \( k^{+} \) would be restricted to low-dimensional applications.

Nonetheless, recent work~\citep{salvi_signature_2021} proved that for two continuously differentiable input paths the complete \emph{signature kernel}, 
\begin{equation}\label{eq:sig_kernel}
    {k}^{\oplus}\colon \parens{\anypath_{t}, \otherpath_{t}} \mapsto \inner{\signature \parens{k_{\anypath}}_{t}}{\signature \parens{k_{\otherpath}}_{t}},
\end{equation}
is the solution of a second-order, hyperbolic partial differential equation (PDE) known as Goursat PDE.
Solving this PDE is a problem of complexity \( O \parens{\sigLength^2 \cdot \sigChannels} \), so still restrictive on the discretisation of the path.
However, by its intrinsic nature, the PDE can be parallelised, turning the complexity into \( O \parens{\sigLength \cdot \sigChannels} \), as long as the GPU is able to accommodate the required number of threads.
Therefore the untruncated signature kernel can be efficiently and parallel computed using state-of-the-art hyperbolic PDE solvers and finite-difference evaluations of the static kernel \( k \).

Hence, we can directly apply \( k^{\oplus} \) in~\cref{eq:svmp_score_func} and we now have a way to properly represent sequential data in feature space, resulting in the final gradient update function:
\begin{equation}\label{eq:sigmp_score_func}
    {\scoreFunc}^{*} (\anypath) = \expectation [\big]
    [k^{\oplus} \parens{\otherpath, \anypath} \grad_{\otherpath} \log \pPdf[\otherpath \given \optimality] + \grad_{\otherpath} k^{\oplus} \parens{\otherpath, \anypath}],
\end{equation}
where the expectation is taken by sampling paths \( \otherpath \) from \( \hat{\qPdf} \).
For convenience, we will use the acronym \sigopt\ whether the algorithm is used for planning or control problems.
A complete overview of the algorithm is presented in~\cref{algo:sigsvgd}.

\section{Results}\label{sec:results}
In this section we present results to demonstrate the correctness and applicability of our method in a set of simulated experiments, ranging from simple 2D motion planning to a challenging benchmark for robotic manipulators.

\subsection{Motion Planning on 2D Terrain}\label{sec:exp_2d_planning}
Our first set of experiments consists of trajectory optimisation in a randomised 2D terrain illustrated in~\cref{fig:2d_planning}.
Regions of higher cost, or hills, are shown in a darker shade whereas valleys are in a lighter colour.
The terrain is parameterised by a series of isotropic Multivariate Gaussian distributions placed randomly according to a Halton sequence and aggregated into a Gaussian Mixture Model denoted by \( \pPdf_{\text{map}} \).

Paths are parameterised by natural cubic splines with \( \splineKnots = 2 \) intermediary knots, apart from the start and goal state.
Our goal is to find the best placement for these knots to find paths from origin to goal that avoid regions of high cost but are not too long.
We adopt the following cost function in order to balance trajectory length and navigability:
\begin{equation}\label{eq:2d_palnning_cost}
\costFn[\state_{t}] = \sum_{t \in \bracks{\llim, \ulim}} \parens[\Big]{\pPdf_{\text{map}}[\state_{t}] + 75 \, \norm{\state_{t} - \state_{t-1}}_{2}},
\end{equation}
where the \( \ell^{2} \)-norm term is a piecewise linear approximation of the trajectory length.
To ensure the approximation is valid each trajectory is decimated into 100 waypoints before being evaluated by~\cref{eq:2d_palnning_cost}.

The initial knots are randomly placed and the plots in~\cref{fig:2d_planning} show the final 20 trajectories found with three different optimisation methods.
Furthermore, the colour of each path depicts its normalised final cost.
On the left we can see the solutions found with Batch Gradient Descent (BGD) and note how all trajectories converge to two modes of similar cost.
The SVMP results are more diverse, but failed to capture one of the BGD modes.
Also note how, when multiple trajectories converge to a single trough, the spline knots are pushed away by the repulsive force resulting in suboptimal solutions.  
On the other hand, the trajectories found by \sigopt\ are not only more diverse, finding more homotopic solutions, but are also able to coexist in the narrow valleys.
This is possible since the repulsive force is being computed in the signature space and not based on the placement of the knots.
Furthermore, notice how for the same reason the paths are more direct and coordinated when compared to SVMP.

\subsection{Point-mass Navigation on an Obstacle Grid}\label{sec:exp_nav}
\begin{table}
    \centering
    \begin{tabular}{lccc}\toprule
         & Cost & Steps \\
        \midrule\midrule
        \sigopt\ & \textbf{1056.0 (58.4)} & \textbf{189.3 (12.6)} \\
        SVMPC & 1396.4 (73.0) & 239.1 (49.4) \\
        MPPI & 1740.7 (192.3) & 290.8 (23.7) \\
        CMA-ES$^{\dag}$ & --- & --- \\
        \bottomrule
    \end{tabular}
    \caption{\label{tab:part2d_results}\textbf{Point-mass navigation results}. 
        The table shows the mean and standard deviation for 20 episodes.
        \emph{Cost} indicates the total accrued cost over the episode. CMA-ES cost is not shown as it couldn't complete the task on any episodes.
        \emph{Steps} indicates the total amount of time-steps the controller needed to reach the goal.
        $^{\dag}$CMA-ES couldn't complete any episodes, so results are omitted.
    }
\end{table}

\begin{figure}
    \centering
    \includegraphics[width=\linewidth]{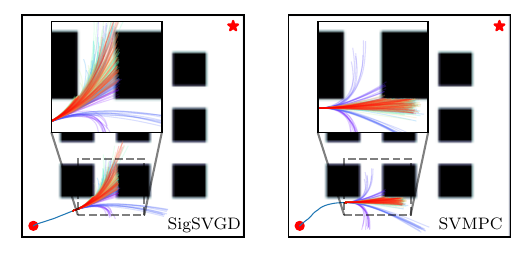}
    \caption{\label{fig:part2d_results}
        \textbf{Point-mass navigation trajectories}.
        The plot shows an intermediate time-step of the navigation task for \sigopt, on the left, and SVMPC, on the right.
        An inset plot enlarges a patch of the map just ahead of the point-mass.
        The rollout colour indicate from which of the policies, \ie\ paths in the optimisation, they originate, whereas fixed motion primitives are shown in purple.
        Note how rollouts generated by \sigopt\ are more disperse, providing a better gradient for policy updates.
        }
\end{figure}

Here, our goal is to demonstrate the benefits of applying the signature kernel Model Predictive Control (MPC).
To that end, we reproduce the point-mass planar navigation task presented in~\citep{barcelos_dual_2021,lambert_stein_2020} and compare SVMPC against and a modified implementation using \sigopt.
The objective is to navigate an holonomic point-mass robot from start to goal through an obstacle grid.
Since the system dynamics is represented as a double integrator model with non-unitary mass $m$, the particle acceleration is given by $\Ddot{\state} = m^{-1} \ctrl$ and the control signal is the force applied to the point-mass.
We adopt the same cost function as in~\citep{barcelos_dual_2021}, that is:
\begin{equation*}
\begin{split}
    &\costFn[\state_{\tIdx}, \ctrl_{\tIdx}] =
    0.5 \, \anyerror_{\tIdx}^\transpose \anyerror_{\tIdx}
    + 0.25 \, \dot{\state_{\tIdx}}^\transpose \dot{\state_{\tIdx}}
    + 0.2 \, \ctrl_{\tIdx}^\transpose \ctrl_{\tIdx}
    + \Bbbone \{\operatorname{col.}\} \, p \\
    &\costFn_{\text{term}}[\state_{\tIdx}, \ctrl_{\tIdx}] =
    1000 \, \anyerror_{\tIdx}^\transpose \anyerror_{\tIdx}
    + 0.1 \, \dot{\state_{\tIdx}}^\transpose \dot{\state_{\tIdx}} \,,
\end{split}
\end{equation*}
where $\anyerror_{\tIdx} = \state_{\tIdx} - \stateGoal$ is the instantaneous position error and $p = 10^6$ is the penalty when a collision happens.

\begin{figure*}
    \centering
    \includegraphics{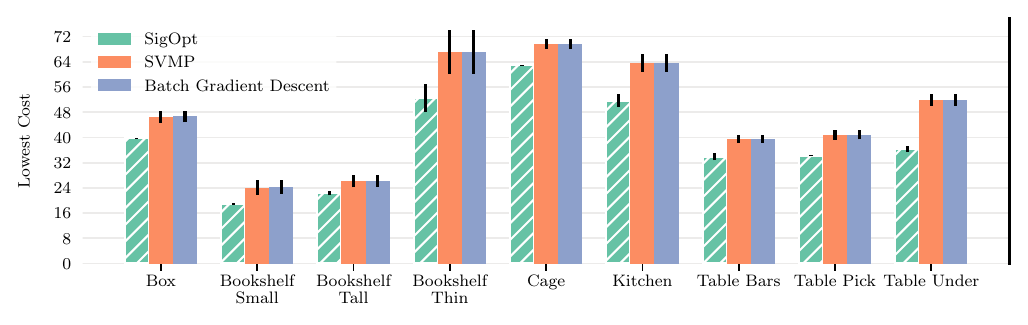}
    \includegraphics{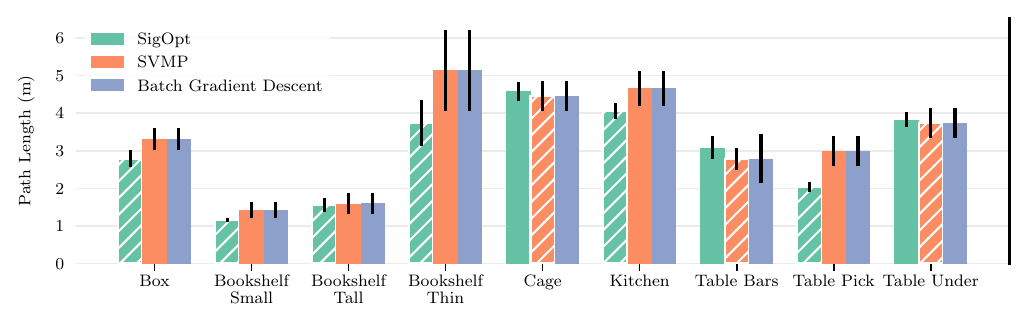}
    \includegraphics{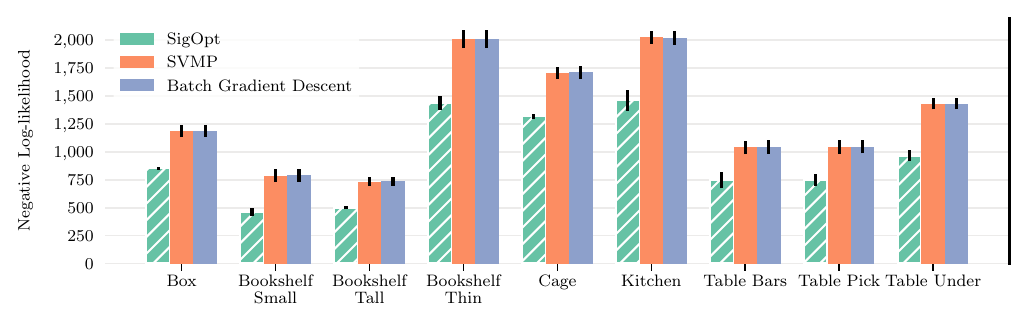}
    \caption{\label{tab:motionbench_results}
        \textbf{Motion planning benchmark}. 
        Results shown are the mean and standard deviation over 5 episodes for 4 distinct requests, totalling 20 iterations per scene.
        Best result is highlighted with a hatched bar.
        \emph{Lowest cost} depicts the cost of the best trajectory found.
        \emph{Path length} is the piecewise linear approximation of the end-effector trajectory length for the best trajectory.
        \emph{NLL} indicates the negative log likelihood and, since we are using an exponential likelihood, represents the total cost of all sampled trajectories.
    }
\end{figure*}

To create a controlled environment with several multi-modal solutions, obstacles are placed equidistantly in a grid (see~\cref{fig:part2d_results}).
The simulator performs a simple collision check based on the particle's state and prevents any future movement in case a collision is detected, simulating a crash.
Barriers are also placed at the environment boundaries to prevent the robot from easily circumventing the obstacle grid.
As the indicator function makes the cost function non-differentiable, we need to compute approximate gradients using Monte Carlo sampling~\citep{lambert_stein_2020}.
Furthermore, since we are using a stochastic controller, we also include CMA-ES and Model Predictive Path Integral (MPPI)~\citep{williams_aggressive_2016} in the benchmark.
A detailed account of the hyper-parameters used in the experiment is presented in \cref{app:exp_hyperparams}.

In this experiment, each of the particles in the optimisation is a path that represents the mean of a stochastic control policy.
Gradients for the policy updates are generated by sampling the control policies and evaluating \emph{rollouts} via an implicit model of the environment.
As CMA-ES only entertains a single solution at any given time, to make the results comparable we increase the amount of samples it evaluates at each step to be equivalent to the number of policies times the number of samples in SVMPC.
One addition to the algorithm in~\citep{lambert_stein_2020} is the inclusion of particles with predefined primitive control policies which are not optimised.
For example, a policy which constantly applies the minimum, maximum, or no acceleration are all valid primitives.
These primitive policies are also included in every candidate solution set of CMA-ES.

The inlay plot in~\cref{fig:part2d_results} illustrates how \sigopt\ promotes policies that are more diverse, covering more of the state-space on forward rollouts.
The outcome can be seen on~\cref{tab:part2d_results}.
\sigopt\ finds lower cost policies and is able to reach the goal in fewer steps than SVMPC.
Due to the dynamical nature of the problem, we are unable to run the optimisation for many iterations during each time-step, as we need to get actions from the controller at a fast rate. 
This poses a challenge to CMA-ES, which crashed on all episodes despite having a much larger number of samples per step.

\subsection{Benchmark Comparison on Robotic Manipulator}

\begin{figure*}
    \centering
    \begin{subfigure}{.25\linewidth}
        \centering
        \includegraphics[width=\linewidth]{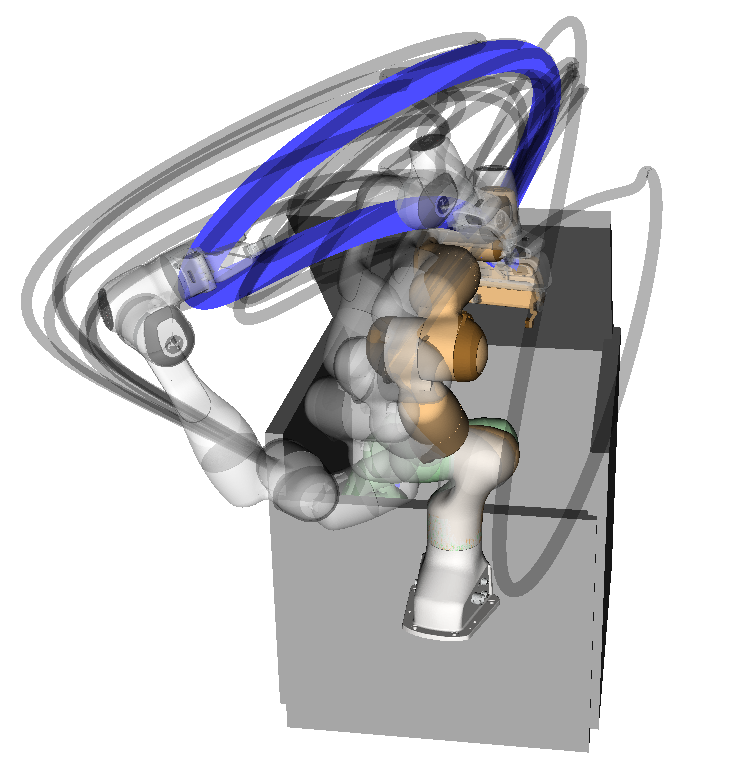}
        \caption{Box}
    \end{subfigure}%
    \begin{subfigure}{.25\linewidth}
        \centering
        \includegraphics[width=\linewidth]{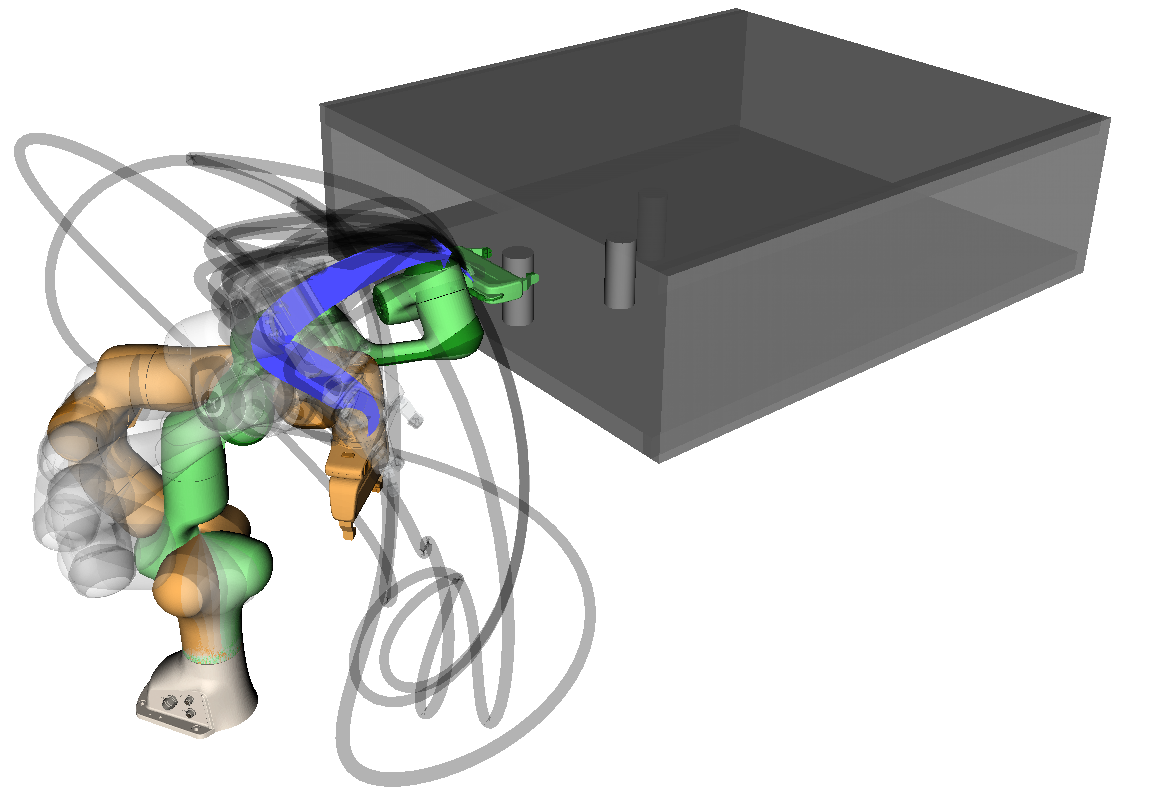}
        \caption{Bookshelf Small}
    \end{subfigure}%
    \begin{subfigure}{.25\linewidth}
        \centering
        \includegraphics[width=\linewidth]{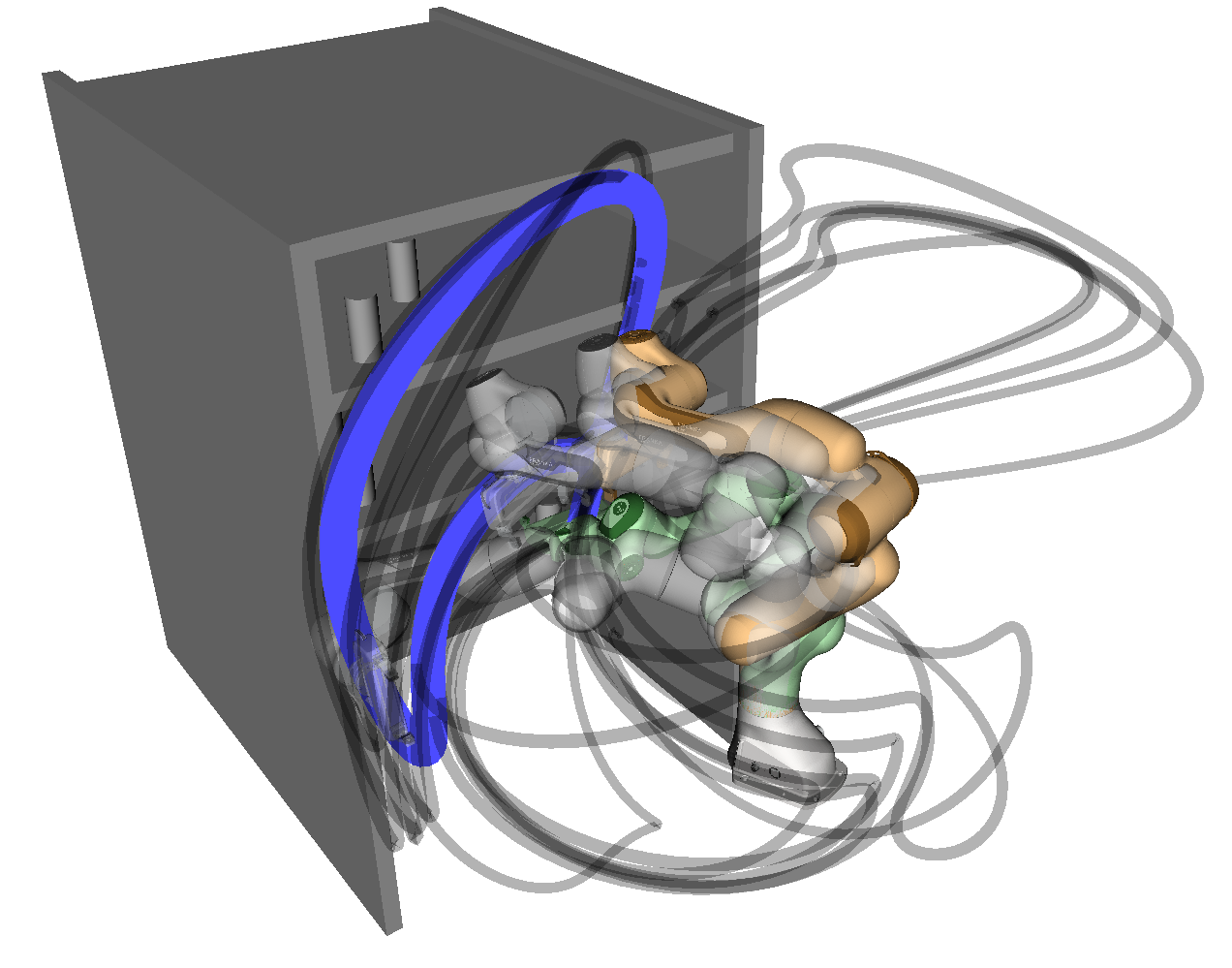}
        \caption{Bookshelf Tall}
    \end{subfigure}%
    \begin{subfigure}{.25\linewidth}
        \centering
        \includegraphics[width=\linewidth]{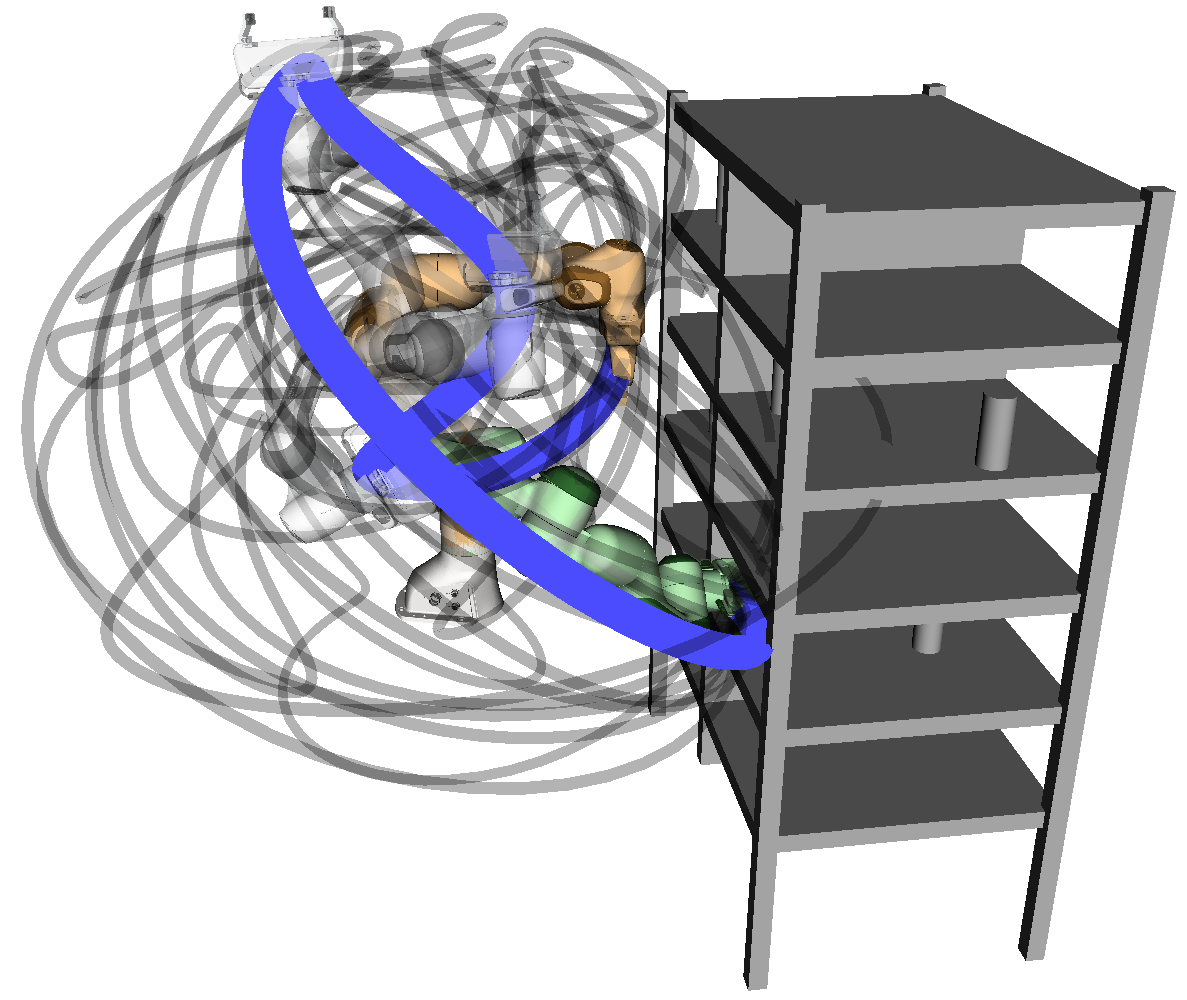}
        \caption{Bookshelf Thin}
    \end{subfigure}
    \begin{subfigure}{.25\linewidth}
        \centering
        \includegraphics[width=\linewidth]{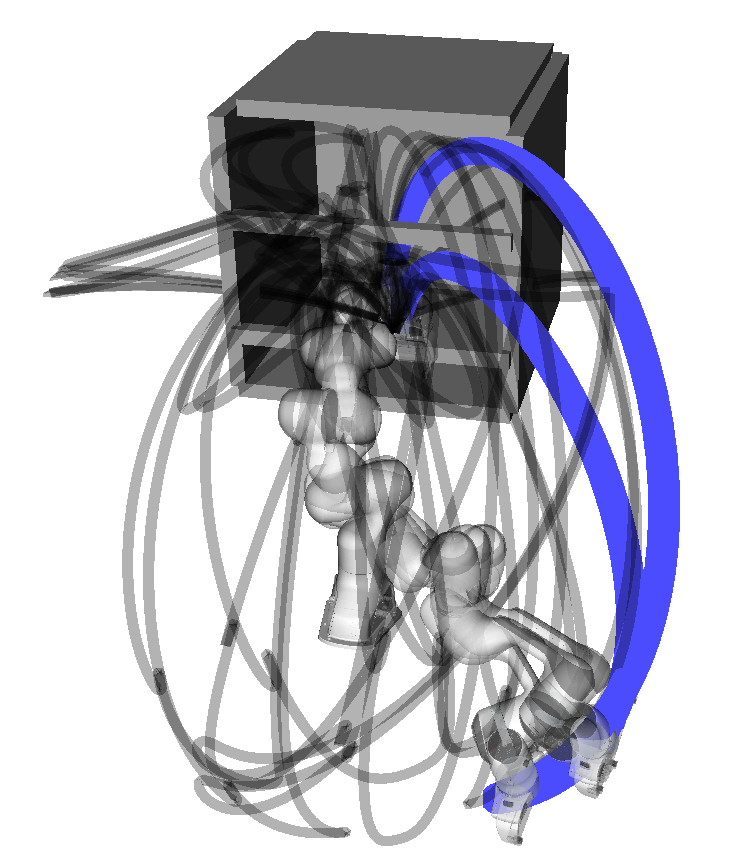}
        \caption{Cage}
    \end{subfigure}%
    \begin{subfigure}{.25\linewidth}
        \centering
        \includegraphics[width=\linewidth]{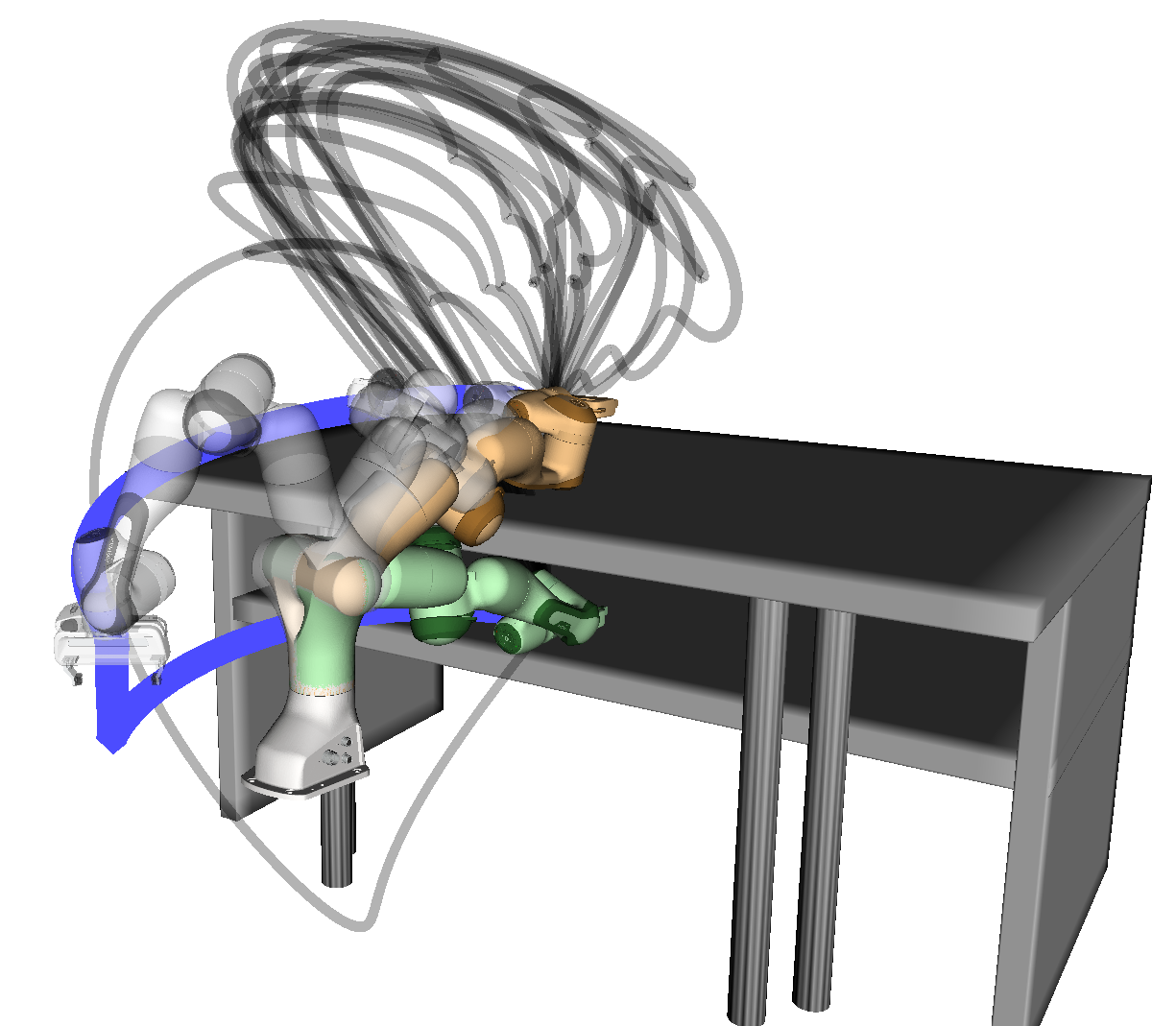}
        \caption{Table Bars}
    \end{subfigure}%
    \begin{subfigure}{.25\linewidth}
        \centering
        \includegraphics[width=\linewidth]{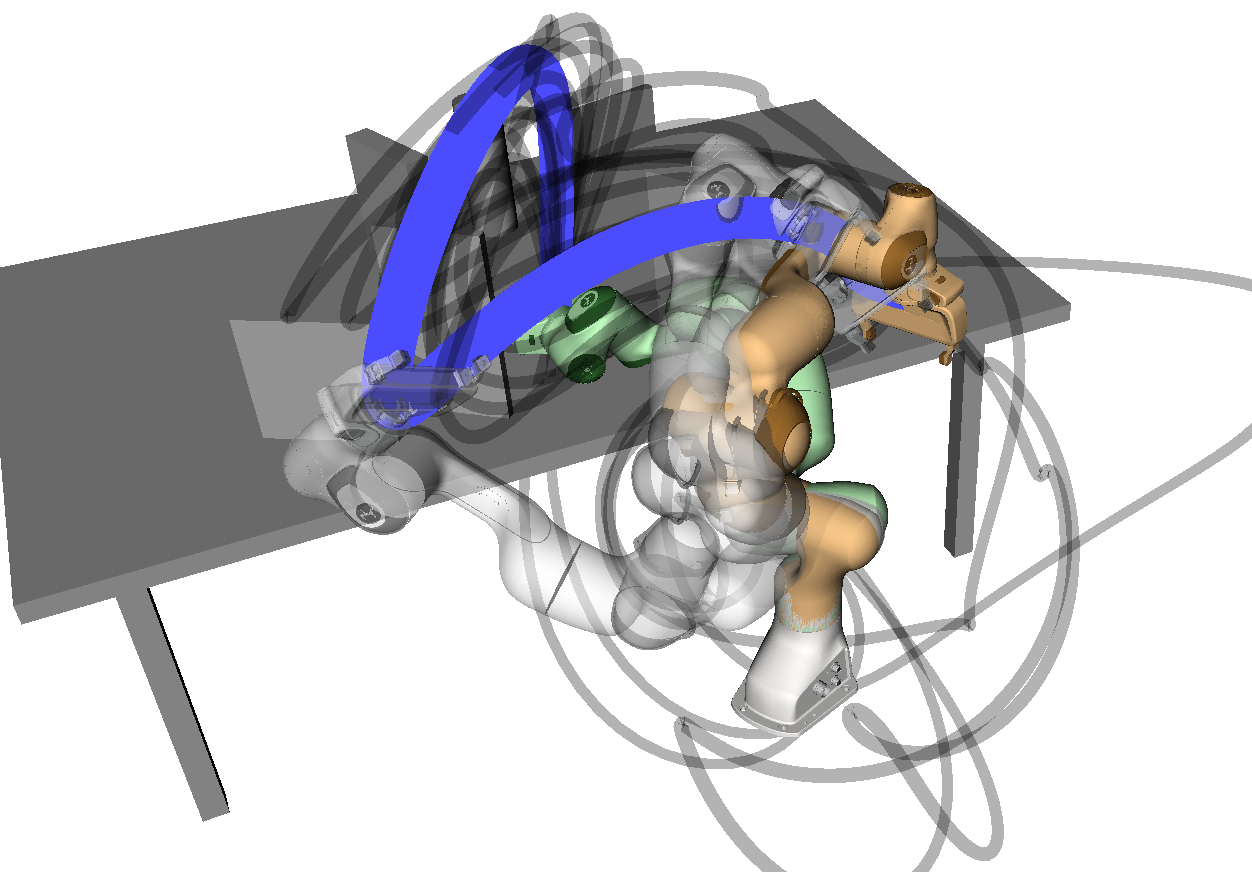}
        \caption{Table Pick}
    \end{subfigure}%
    \begin{subfigure}{.25\linewidth}
        \centering
        \includegraphics[width=\linewidth]{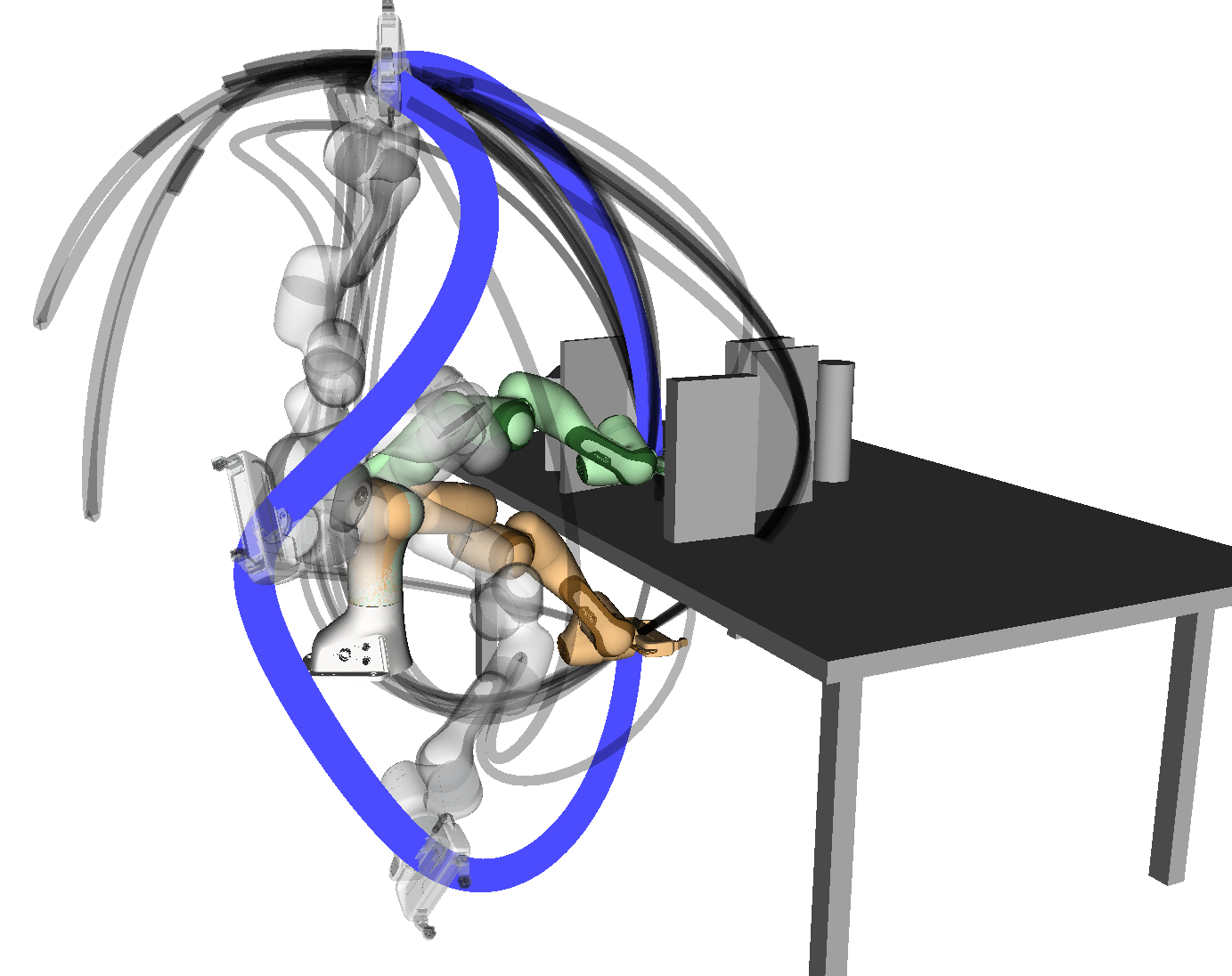}
        \caption{Table Under}
    \end{subfigure}
    \caption{\label{fig:motionbench_paths}
        \textbf{Visualisation of \sigopt\ in the motion planning benchmark.}
        The \emph{Blue} and \emph{Grey} lines denote the end-effector's trajectories with the former highlighting the trajectory with the lowest cost.
        The \emph{Orange} and \emph{Green} tinted robot poses denote the start and target configuration, respectively.
        The translucent robot poses denote in-between configurations of the lowest-cost solution.
    }
\end{figure*}

To test our approach on a more complex planning problem we compare batch gradient descent (i.e. parallel gradient descent on different initialisations), SVMP and \sigopt\ in robotic manipulation problems generated using MotionBenchMaker~\citep{chamzas_motionbenchmaker_2022}.
A problem consists of a scene with randomly placed obstacles and a consistent request to move the manipulator from its starting pose to a target configuration.
For each scene in the benchmark, we generate 4 different requests and run the optimisation with 5 random seeds for a total of 20 episodes per scene.

The robot used is a Franka Emika Panda with 7 Degrees of Freedom (DOF).
The cost function is designed to generate trajectories that are smooth, collision-free and with a short displacement of the robot's end-effector.
We once again resort to a fully-differentiable function to reduce the extraneous influence of approximating gradients with Monte Carlo samples.
As is typical in motion planning, the optimisation is performed directly in \emph{configuration space} (C-space), which simplifies the search for feasible plans.
To reduce the sampling space and promote smooth trajectories, we once again parameterise the path of each of the robot joints with natural cubic splines, adopting 3 intermediary knots besides those at the initial and target poses.

\subsubsection{Regularising Path Length and Dynamical Motions}
\hfill\\
Finally, the use of splines to interpolate the trajectories ensures smoothness in generated trajectories, but that does not necessarily imply in smooth dynamics for the manipulator.
To visualise this, consider, for example, a trajectory in \( \cSpace \) parameterised by a natural cubic spline.
The configurations \( \cState \) in between each knot can be interpolated, resulting in a smooth trajectory of the robot end-effector in Euclidean coordinates in SE(3).
However, the same end-effector trajectory could be traversed in a constant linear speed or with a jerky acceleration and deceleration motion.
More specifically, if we use a fixed number of interpolated configurations between knots without care to impose dynamical restrictions to the simulator, knots that are further apart will result in motions with greater speed and acceleration since a larger distance would be covered during the same interval.
To avoid these abrupt motions on the robots joints, we introduce the term \( \dynCost \) to the cost function, which penalises the linear distance between consecutive configurations:
\begin{equation}\label{eq:dyn_cost}
    \dynCost = \sum_{\primIdx = 2}^{\interPts} \costWeight^{\transpose} \norm{\cState_{\primIdx} - \cState_{\primIdx - 1}}_{2},
\end{equation}
where \( \interPts \) is the number of intermediary configurations chosen when discretising the path spline and the weight \( \costWeight \) can be used to assign a higher importance to certain robot joints.
We choose to adopt a vector \( \costWeight \) which is a linear interpolation from 1 to 0.7, where the higher value is assigned to the base joint of the manipulator and progressively reduced until the end-effector.
A similar approach as the one presented in~\cref{eq:dyn_cost} can be used to penalise the length of the robot's trajectory in workspace.
We include a final term to our cost function, \( \lenCost \), that penalises exclusively the length of the end-effector path. 
This brings us to our final cost function:
\begin{equation}\label{eq:manipulator_total_cost}
    \costFn = 2.5 \, \lenCost + 2.5 \, \dynCost + \colCost + 10 \, \scolCost,
\end{equation}
where each of the terms are respectively the cost for path length, path dynamics, collision with the environment and self-collision.
The optimisation is carried out for 500 iterations and the kernel repulsive force is scheduled with cosine annealing~\citep{loshchilov_sgdr_2017}.
By reducing the repulsive force on the last portion of the optimisation, we allow trajectories at the same local minima to converge to the modes and are able to qualitatively measure the diversity of each approach.

The results shown on~\cref{tab:motionbench_results} demonstrate how \sigopt\ achieves better results in almost all metrics for every scenario.
The proper representation of paths results in better exploration of the configuration space and leads to better global properties of the solutions found.
This can be seen in~\cref{fig:motionbench_paths}, which shows the end-effector paths for \sigopt\ and SVMP.
One of such paths is also illustrated in~\cref{fig:intro-image}.
Results found by \sigopt\ also show a higher percentage of feasible trajectories and lower contact depths for rollouts in collision (see~\cref{tab:sim_col_results}).

\begin{table*}[ht]
    \centering
    \begin{tabular}{lcccccc}\toprule
         & \multicolumn{2}{c}{\sigopt} & \multicolumn{2}{c}{SVMP} & \multicolumn{2}{c}{Batch Gradient Descent}\\
        \cmidrule(lr){2-3}\cmidrule(lr){4-5}\cmidrule(lr){6-7}
        Scene        & Contact Depth & Feasible Pct. & Contact Depth & Feasible Pct. & Contact Depth & Feasible Pct. \\
        \midrule\midrule
Box             & 3.74 (2.30)          & \textbf{94.99 (3.78)} & \textbf{3.62 (1.95)} & 94.96 (3.32) & 3.63 (1.95)          & 94.97 (3.31)          \\
Bookshelf Small & \textbf{1.32 (2.50)} & \textbf{96.63 (5.48)} & 1.55 (2.19)          & 96.20 (4.68) & 1.56 (2.20)          & 96.18 (4.71)          \\
Bookshelf Tall  & 0.56 (1.78)          & 98.30 (4.65)          & 0.27 (0.60)          & 99.02 (1.76) & \textbf{0.27 (0.59)} & \textbf{99.03 (1.74)} \\
Bookshelf Thin  & \textbf{2.78 (3.11)} & \textbf{94.59 (4.94)} & 3.14 (3.50)          & 93.54 (5.57) & 3.14 (3.50)          & 93.54 (5.57)          \\
Cage            & 2.13 (1.82)          & \textbf{96.12 (2.92)} & \textbf{2.00 (1.67)} & 96.11 (2.89) & \textbf{2.00 (1.67)} & 96.11 (2.89)          \\
Kitchen         & \textbf{9.82 (6.95)} & 88.04 (9.85)          & 10.61 (6.45)         & 88.59 (6.21) & 10.62 (6.71)         & \textbf{88.61 (6.21)} \\
Table Bars      & \textbf{9.46 (7.43)} & \textbf{92.42 (5.89)} & 9.52 (8.05)          & 92.09 (6.69) & 9.70 (8.44)          & 92.05 (6.85)          \\
Table Pick      & \textbf{0.22 (0.67)} & \textbf{99.56 (1.67)} & 0.83 (1.04)          & 98.06 (2.62) & 0.83 (1.02)          & 98.08 (2.43)          \\
Table Under     & \textbf{3.33 (2.60)} & \textbf{93.63 (5.36)} & 5.16 (4.75)          & 90.19 (8.21) & 5.18 (4.77)          & 90.06 (8.30)         \\
        \bottomrule
    \end{tabular}
    \caption{\label{tab:sim_col_results}
        \textbf{Motion planning benchmark}. 
        Results shown are the mean and standard deviation over 5 episodes for 4 distinct requests, totalling 20 iterations per scene.
        \emph{Contact Depth} indicates the average collision depth of the trajectories found (in millimetres), if a collision happens.
        \emph{Feasible Pct.} is the average percentage of the trajectory that is collision-free.
    }
\end{table*}

\subsubsection{Robot Collision as Continuous Cost}
\hfill\\
Typically collision-checking is a binary check and non-differentiable.
To generate differentiable collision checking with informative gradients, we resort to continuous occupancy grids.
Occupancy grid maps are often generated from noisy and uncertain sensor measurement by discretising the space $\wSpace$ where the robot operates (know as \emph{workspace}) into grid-cells, where each cell represents an evenly spaced field of binary random variables that corresponds to the presence of an obstacle at the given location.
However, the discontinuity in-between each cell means these grid maps are non-differentiable and not suitable for optimisation-based planning.
A continuous analogue of an occupancy map can be generalised by a kernelised projection to high-dimensional spaces~\citep{ramos_hilbert_2016} or with distance-based methods~\citep{jones_3d_2006}.

In this work we trade off the extra complexity of the methods previously mentioned for a coarser but simpler approach.
Inspired by~\citep{danielczuk_object_2021}, we learn the occupancy of each scene using a neural network as a universal function approximator.
We train the network to approximate a continuous function that returns the likelihood of a robot configuration being occupied.
The rationale for this choice is that, since all methods are optimised under the same conditions, the comparative results should not be substantially impacted by the overall quality of the map.
Additionally, the trained network is fast to query and fast to obtain derivatives with respect to inputs, properties that are beneficial for querying of large batches of coordinates for motion planning. 

Given a dataset of $\dataSize$ pairs of coordinates and a binary value which indicates whether the coordinate is occupied, \ie\ $\dataset = \set{ \parens{\wState_{\primIdx}, \binresp_{\primIdx}} }_{\primIdx=1}^{\dataSize}$, where $\wState_{\primIdx} \in \wSpace \subseteq \R^{\wDim}$, and $\binresp_{\primIdx} \in \set{0,1}$, for $\primIdx=1,\ldots, \dataSize$.
The network then learns a mapping $\fcol$ between a coordinate of interest $\preds$ and the probability of it being occupied, that is, $\fcol[\wState] = \prob[\binresp = 1 \given \preds]$.
A dataset of this format can be obtained, for instance, from depth sensors as point clouds.
We model $\fcol$ as a fully-connected neural network, with $\tanh$ as the activation function between hidden layers, and $\mathrm{sigmoid}$ as the output layer.
The final network is akin to a binary classification problem, which can be learned via a binary cross-entropy loss with gradient descent optimisers.
As such, we can construct a collision cost function $\fcol\colon \wSpace \to \R$ that maps workspace coordinates into cost values associated at the corresponding locations.

A similar problem occurs when ascertaining whether a given configuration of the robot's joints is unfeasible, leading to a self-collision.
We address this issue in a similar manner, by training a separate neural network to approximate a continuous function \( \fscol \) which maps configurations of the robot to the likelihood of they being in self-collision.
More precisely, \( \fscol\colon \cSpace \to \R \), where \(  \fscol[\cState] = \prob[\binresp = 1 \given \cState] \), for $\cState_{\primIdx} \in \cSpace \subseteq \R^{\cDim}$, and $\binresp_{\primIdx} \in \set{0,1}$.
The dataset used to train \( \fscol \) is generated by randomly choosing configurations within the joint limits of the robot and performing a binary self-collision check provided by the robot's API.

\subsubsection{Bringing Collision Cost from Workspace to Configuration Space}
\hfill\\
Collision checking requires information about the workspace geometry of the robot to determine whether it overlaps with objects in the environment. On the other hand, we assume that the robot movement is defined and optimised in C-space.
The cost functions to shape robot behaviour are often defined in the Cartesian task space.
We denote C-space as $\cSpace \subseteq \R^{\cDim}$, where there are $\cDim$ joints in the case of a robotic manipulator.
The joint configurations, $\cState \in \cSpace$, are elements of the C-space, while Cartesian coordinates in task space are denoted as $\wState \in \wSpace$.
We now outline the procedure of \emph{pulling} a cost gradient defined in the workspace to the C-space.

We start by defining $\bodyPts$ body points on the robot, each with a forward kinematics function $\ffk_{i}$ mapping configurations to the Cartesian coordinates $\wState_i$ at the body point, $\ffk_{i} \colon \cSpace \to \wSpace$, for each $\primIdx = 1, \ldots, \bodyPts$.
Let the Jacobian of the forward kinematics functions \wrt\ the joint configurations be denoted as
\begin{equation}
    {\jacob[\cdot]}_{\ffk}^{\primIdx} = \frac{\dif\ffk_{\primIdx}}{\dif\cState} \parens{\cdot}.
\end{equation}
The derivative of a cost potential $\colCost$ which operates on the body points, such as the occupancy cost potential, can then be \emph{pulled} into the C-space with:
\begin{equation}
    \grad_{\cState} \costFn = \sum_{\primIdx = 1}^{\bodyPts} {\jacob[\cState]}_{\ffk}^{\primIdx} \grad_{\wState} \costFn,
\end{equation}
which allows us to update trajectory in the C-space $\cSpace$ with cost in the Cartesian space $\wSpace$.

\section{Conclusion}\label{sec:conclusion}
This work, to the best of our knowledge, is the first to introduce the use of path signatures for trajectory optimisation in robotics.
We discuss how this transformation can be used as a canonical \emph{linear} feature map to represent trajectories and how it possesses many desirable properties, such as invariance under time reparametrisation.
We use these ideas to construct \sigopt, a kernel method to solve control and motion planning problems in a variational inference setting.
It approximates the posterior distribution over optimal paths with an empirical distribution comprised of a set of vector-valued particles which are all optimised in parallel.

In previous work it has been shown that approaching the optimisation from the variational perspective alleviates the problem of local optimality, providing a more diverse set of solutions.
We argue that the use of signatures improves on previous work and can lead to even better global properties.
Despite the signature poor scalability, we show how we can construct fast and paralellisable signature kernels by leveraging recent results in rough path theory.
The RKHS induced by this kernel creates a structured space that captures the sequential nature of paths.
This is demonstrated through an extensive set of experiments that the structure provided helps the functional optimisation, leading to better global solutions than equivalent methods without it.
We hope the ideas herein presented will serve an inspiration for further research and stimulate a groundswell of new work capitalising on the benefits of signatures in many other fields within the robotics community.

\balance{}
\printbibliography{}

@article{jaillet_path_2008,
	title = {Path {Deformation} {Roadmaps}: {Compact} {Graphs} with {Useful} {Cycles} for {Motion} {Planning}},
	volume = {27},
	issn = {0278-3649},
	shorttitle = {Path {Deformation} {Roadmaps}},
	url = {https://doi.org/10.1177/0278364908098411},
	doi = {10.1177/0278364908098411},
	language = {en},
	number = {11-12},
	urldate = {2023-06-13},
	journal = {The International Journal of Robotics Research},
	author = {Jaillet, Léonard and Simeon, Thierry},
	month = nov,
	year = {2008},
	note = {Publisher: SAGE Publications Ltd STM},
	pages = {1175--1188},
}

@book{camacho_model_2013,
	address = {London},
	edition = {2},
	series = {Advanced {Textbooks} in {Control} and {Signal} {Processing}},
	title = {Model {Predictive} {Control}},
	isbn = {978-0-85729-398-5},
	url = {https://link.springer.com/book/10.1007/978-0-85729-398-5},
	language = {en},
	publisher = {Springer-Verlag},
	author = {Camacho, Eduardo F. and Alba, Carlos Bordons},
	year = {2013},
}

@book{scholkopf_learning_2002,
	address = {Cambridge, Mass.},
	edition = {Reprint.},
	series = {Adaptive computation and machine learning series},
	title = {Learning with kernels: support vector machines, regularization, optimization, and beyond},
	isbn = {978-0-262-19475-4 978-0-262-53657-8},
	shorttitle = {Learning with kernels},
	language = {eng},
	publisher = {MIT Press},
	author = {Schölkopf, Bernhard and Smola, Alexander Johannes},
	year = {2002},
}

@article{mukadam_continuous-time_2018,
	title = {Continuous-time {Gaussian} process motion planning via probabilistic inference},
	volume = {37},
	issn = {0278-3649, 1741-3176},
	url = {http://journals.sagepub.com/doi/10.1177/0278364918790369},
	doi = {10.1177/0278364918790369},
	language = {en},
	number = {11},
	urldate = {2022-11-02},
	journal = {The International Journal of Robotics Research},
	author = {Mukadam, Mustafa and Dong, Jing and Yan, Xinyan and Dellaert, Frank and Boots, Byron},
	month = sep,
	year = {2018},
	pages = {1319--1340},
}

@article{lavalle_randomized_2001,
	title = {Randomized {Kinodynamic} {Planning}},
	volume = {20},
	issn = {0278-3649, 1741-3176},
	url = {http://journals.sagepub.com/doi/10.1177/02783640122067453},
	doi = {10.1177/02783640122067453},
	language = {en},
	number = {5},
	urldate = {2022-11-02},
	journal = {The International Journal of Robotics Research},
	author = {LaValle, Steven M. and Kuffner, James J.},
	month = may,
	year = {2001},
	pages = {378--400},
}

@inproceedings{marinho_functional_2016,
	title = {Functional {Gradient} {Motion} {Planning} in {Reproducing} {Kernel} {Hilbert} {Spaces}},
	isbn = {978-0-9923747-2-3},
	url = {http://www.roboticsproceedings.org/rss12/p46.pdf},
	doi = {10.15607/RSS.2016.XII.046},
	language = {en},
	urldate = {2023-01-10},
	booktitle = {Robotics: {Science} and {Systems} {XII}},
	publisher = {Robotics: Science and Systems Foundation},
	author = {Marinho, Zita and Boots, Byron and Dragan, Anca and Byravan, Arunkumar and J. Gordon, Geoffrey and Srinivasa, Siddhartha},
	year = {2016},
}

@inproceedings{barcelos_dual_2021,
	title = {Dual online stein variational inference for control and dynamics},
	copyright = {All rights reserved},
	url = {https://doi.org/10.15607/RSS.2021.XVII.068},
	doi = {10.15607/RSS.2021.XVII.068},
	booktitle = {Robotics: {Science} and systems {XVII}, virtual event, july 12-16, 2021},
	author = {Barcelos, Lucas and Lambert, Alexander and Oliveira, Rafael and Borges, Paulo and Boots, Byron and Ramos, Fabio},
	editor = {Shell, Dylan A. and Toussaint, Marc and Hsieh, M. Ani},
	year = {2021},
	note = {tex.bibsource: dblp computer science bibliography, https://dblp.org
tex.biburl: https://dblp.org/rec/conf/rss/BarcelosL0BBR21.bib
tex.timestamp: Wed, 21 Jul 2021 17:07:40 +0200},
}

@article{yu_gaussian_2022,
	title = {A {Gaussian} variational inference approach to motion planning},
	volume = {abs/2209.05655},
	url = {https://doi.org/10.48550/arXiv.2209.05655},
	doi = {10.48550/arXiv.2209.05655},
	urldate = {2023-01-11},
	journal = {Computing Research Repository},
	author = {Yu, Hongzhe and Chen, Yongxin},
	year = {2022},
}

@article{tjanaka_training_2022,
	title = {Training {Diverse} {High}-{Dimensional} {Controllers} by {Scaling} {Matrix} {Adaptation} {MAP}-{Annealing}},
	volume = {abs/2210.02622},
	url = {https://doi.org/10.48550/arXiv.2210.02622},
	doi = {10.48550/arXiv.2210.02622},
	language = {en},
	journal = {Computing Research Repository},
	author = {Tjanaka, Bryon and Fontaine, Matthew C and Kalkar, Aniruddha and Nikolaidis, Stefanos},
	year = {2022},
}

@book{rasmussen_gaussian_2006,
	address = {Cambridge, Mass},
	series = {Adaptive computation and machine learning},
	title = {Gaussian processes for machine learning},
	isbn = {978-0-262-18253-9},
	publisher = {MIT Press},
	author = {Rasmussen, Carl Edward and Williams, Christopher K. I.},
	year = {2006},
	note = {OCLC: ocm61285753},
}

@inproceedings{loshchilov_sgdr_2017,
	address = {Toulon, France},
	title = {{SGDR}: {Stochastic} {Gradient} {Descent} with {Warm} {Restarts}},
	shorttitle = {{SGDR}},
	url = {https://openreview.net/forum?id=Skq89Scxx},
	language = {en},
	urldate = {2023-01-23},
	booktitle = {5th {International} {Conference} on {Learning} {Representations}, \{{ICLR}\} 2017, {Toulon}, {France}, {April} 24-26, 2017, {Conference} {Track} {Proceedings}},
	publisher = {OpenReview.net},
	author = {Loshchilov, Ilya and Hutter, Frank},
	year = {2017},
}

@article{kiraly_kernels_2019,
	title = {Kernels for {Sequentially} {Ordered} {Data}},
	volume = {20},
	url = {http://jmlr.org/papers/v20/16-314.html},
	language = {en},
	number = {31},
	journal = {Journal of Machine Learning Research},
	author = {Kiraly, Franz J and Oberhauser, Harald},
	year = {2019},
	pages = {31:1--31:45},
}

@article{hansen_cma_2016,
	title = {The {CMA} {Evolution} {Strategy}: {A} {Tutorial}},
	shorttitle = {The {CMA} {Evolution} {Strategy}},
	url = {https://arxiv.org/abs/1604.00772v1},
	doi = {10.48550/arXiv.1604.00772},
	language = {en},
	urldate = {2023-01-23},
	journal = {Computing Research Repository},
	author = {Hansen, Nikolaus},
	month = apr,
	year = {2016},
}

@misc{danielczuk_object_2021,
	title = {Object {Rearrangement} {Using} {Learned} {Implicit} {Collision} {Functions}},
	url = {http://arxiv.org/abs/2011.10726},
	urldate = {2023-01-31},
	publisher = {arXiv},
	author = {Danielczuk, Michael and Mousavian, Arsalan and Eppner, Clemens and Fox, Dieter},
	month = mar,
	year = {2021},
	note = {arXiv:2011.10726 [cs]},
}

@article{jones_3d_2006,
	title = {{3D} distance fields: a survey of techniques and applications},
	volume = {12},
	issn = {1941-0506},
	shorttitle = {{3D} distance fields},
	doi = {10.1109/TVCG.2006.56},
	number = {4},
	journal = {IEEE Transactions on Visualization and Computer Graphics},
	author = {Jones, M.W. and Baerentzen, J.A. and Sramek, M.},
	month = jul,
	year = {2006},
	note = {Conference Name: IEEE Transactions on Visualization and Computer Graphics},
	pages = {581--599},
}

@inproceedings{hamalainen_ppo-cma_2020,
	title = {{PPO}-{CMA}: {Proximal} {Policy} {Optimization} with {Covariance} {Matrix} {Adaptation}},
	shorttitle = {{PPO}-{CMA}},
	doi = {10.1109/MLSP49062.2020.9231618},
	booktitle = {2020 {IEEE} 30th {International} {Workshop} on {Machine} {Learning} for {Signal} {Processing} ({MLSP})},
	author = {Hämäläinen, Perttu and Babadi, Amin and Ma, Xiaoxiao and Lehtinen, Jaakko},
	month = sep,
	year = {2020},
	note = {ISSN: 1551-2541},
	pages = {1--6},
}

@article{hansen_reducing_2003,
	title = {Reducing the {Time} {Complexity} of the {Derandomized} {Evolution} {Strategy} with {Covariance} {Matrix} {Adaptation} ({CMA}-{ES})},
	volume = {11},
	issn = {1063-6560, 1530-9304},
	url = {https://direct.mit.edu/evco/article/11/1/1-18/1139},
	doi = {10.1162/106365603321828970},
	language = {en},
	number = {1},
	urldate = {2023-01-23},
	journal = {Evolutionary Computation},
	author = {Hansen, Nikolaus and Müller, Sibylle D. and Koumoutsakos, Petros},
	month = mar,
	year = {2003},
	pages = {1--18},
}

@inproceedings{byravan_space-time_2014,
	address = {Hong Kong, China},
	title = {Space-time functional gradient optimization for motion planning},
	isbn = {978-1-4799-3685-4},
	url = {http://ieeexplore.ieee.org/document/6907818/},
	doi = {10.1109/ICRA.2014.6907818},
	language = {en},
	urldate = {2023-01-10},
	booktitle = {2014 {IEEE} {International} {Conference} on {Robotics} and {Automation} ({ICRA})},
	publisher = {IEEE},
	author = {Byravan, Arunkumar and Boots, Byron and Srinivasa, Siddhartha S. and Fox, Dieter},
	month = may,
	year = {2014},
	pages = {6499--6506},
}

@inproceedings{schulman_finding_2013,
	title = {Finding {Locally} {Optimal}, {Collision}-{Free} {Trajectories} with {Sequential} {Convex} {Optimization}},
	isbn = {978-981-07-3937-9},
	url = {http://www.roboticsproceedings.org/rss09/p31.pdf},
	doi = {10.15607/RSS.2013.IX.031},
	urldate = {2022-12-22},
	booktitle = {Robotics: {Science} and {Systems} {IX}},
	publisher = {Robotics: Science and Systems Foundation},
	author = {Schulman, John and Ho, Jonathan and Lee, Alex and Awwal, Ibrahim and Bradlow, Henry and Abbeel, Pieter},
	month = jun,
	year = {2013},
}

@article{gonzalez_review_2016,
	title = {A {Review} of {Motion} {Planning} {Techniques} for {Automated} {Vehicles}},
	volume = {17},
	issn = {1524-9050, 1558-0016},
	url = {http://ieeexplore.ieee.org/document/7339478/},
	doi = {10.1109/TITS.2015.2498841},
	number = {4},
	urldate = {2022-12-22},
	journal = {IEEE Transactions on Intelligent Transportation Systems},
	author = {Gonzalez, David and Perez, Joshue and Milanes, Vicente and Nashashibi, Fawzi},
	month = apr,
	year = {2016},
	pages = {1135--1145},
}

@article{berntorp_models_2014,
	title = {Models and methodology for optimal trajectory generation in safety-critical road–vehicle manoeuvres},
	volume = {52},
	issn = {0042-3114, 1744-5159},
	url = {http://www.tandfonline.com/doi/abs/10.1080/00423114.2014.939094},
	doi = {10.1080/00423114.2014.939094},
	language = {en},
	number = {10},
	urldate = {2022-12-22},
	journal = {Vehicle System Dynamics},
	author = {Berntorp, Karl and Olofsson, Björn and Lundahl, Kristoffer and Nielsen, Lars},
	month = oct,
	year = {2014},
	pages = {1304--1332},
}

@article{heilmeier_minimum_2020,
	title = {Minimum curvature trajectory planning and control for an autonomous race car},
	volume = {58},
	issn = {0042-3114, 1744-5159},
	url = {https://www.tandfonline.com/doi/full/10.1080/00423114.2019.1631455},
	doi = {10.1080/00423114.2019.1631455},
	language = {en},
	number = {10},
	urldate = {2022-12-22},
	journal = {Vehicle System Dynamics},
	author = {Heilmeier, Alexander and Wischnewski, Alexander and Hermansdorfer, Leonhard and Betz, Johannes and Lienkamp, Markus and Lohmann, Boris},
	month = oct,
	year = {2020},
	pages = {1497--1527},
}

@article{kavraki_probabilistic_1996,
	title = {Probabilistic roadmaps for path planning in high-dimensional configuration spaces},
	volume = {12},
	issn = {1042296X},
	url = {http://ieeexplore.ieee.org/document/508439/},
	doi = {10.1109/70.508439},
	number = {4},
	urldate = {2022-12-22},
	journal = {IEEE Transactions on Robotics and Automation},
	author = {Kavraki, L.E. and Svestka, P. and Latombe, J.-C. and Overmars, M.H.},
	month = aug,
	year = {1996},
	pages = {566--580},
}

@article{al-bluwi_motion_2012,
	title = {Motion planning algorithms for molecular simulations: {A} survey},
	volume = {6},
	issn = {15740137},
	shorttitle = {Motion planning algorithms for molecular simulations},
	url = {https://linkinghub.elsevier.com/retrieve/pii/S157401371200024X},
	doi = {10.1016/j.cosrev.2012.07.002},
	language = {en},
	number = {4},
	urldate = {2022-12-22},
	journal = {Computer Science Review},
	author = {Al-Bluwi, Ibrahim and Siméon, Thierry and Cortés, Juan},
	month = jul,
	year = {2012},
	pages = {125--143},
}

@article{gammell_asymptotically_2021,
	title = {Asymptotically {Optimal} {Sampling}-{Based} {Motion} {Planning} {Methods}},
	volume = {4},
	issn = {2573-5144, 2573-5144},
	url = {https://www.annualreviews.org/doi/10.1146/annurev-control-061920-093753},
	doi = {10.1146/annurev-control-061920-093753},
	language = {en},
	number = {1},
	urldate = {2022-12-22},
	journal = {Annual Review of Control, Robotics, and Autonomous Systems},
	author = {Gammell, Jonathan D. and Strub, Marlin P.},
	month = may,
	year = {2021},
	pages = {295--318},
}

@book{lavalle_planning_2006,
	address = {Cambridge ; New York},
	title = {Planning algorithms},
	isbn = {978-0-521-86205-9},
	publisher = {Cambridge University Press},
	author = {LaValle, Steven Michael},
	year = {2006},
	note = {OCLC: ocm65301992},
}

@article{haugh_tutorial_2021,
	title = {A {Tutorial} on {Markov} {Chain} {Monte}-{Carlo} and {Bayesian} {Modeling}},
	issn = {1556-5068},
	url = {https://www.ssrn.com/abstract=3759243},
	doi = {10.2139/ssrn.3759243},
	language = {en},
	urldate = {2022-12-02},
	journal = {SSRN Electronic Journal},
	author = {Haugh, Martin B.},
	year = {2021},
}

@inproceedings{dong_motion_2016,
	title = {Motion {Planning} as {Probabilistic} {Inference} using {Gaussian} {Processes} and {Factor} {Graphs}},
	isbn = {978-0-9923747-2-3},
	url = {http://www.roboticsproceedings.org/rss12/p01.pdf},
	doi = {10.15607/RSS.2016.XII.001},
	language = {en},
	urldate = {2022-12-01},
	booktitle = {Robotics: {Science} and {Systems} {XII}},
	publisher = {Robotics: Science and Systems Foundation},
	author = {Dong, Jing and Mukadam, Mustafa and Dellaert, Frank and Boots, Byron},
	year = {2016},
}

@book{lyons_differential_2007,
	address = {Berlin New York},
	series = {Lecture notes in mathematics},
	title = {Differential equations driven by rough paths: École d'ete de probabilites de {Saint}-{Flour} {XXXIV}-2004},
	isbn = {978-3-540-71285-5},
	shorttitle = {Differential equations driven by rough paths},
	language = {eng},
	number = {1908},
	publisher = {Springer},
	author = {Lyons, Terry J. and Picard, Jean and Lévy, Thierry},
	collaborator = {Ecole d'été de probabilités de Saint-Flour},
	year = {2007},
}

@misc{yang_developing_2017,
	title = {Developing the {Path} {Signature} {Methodology} and its {Application} to {Landmark}-based {Human} {Action} {Recognition}},
	url = {http://arxiv.org/abs/1707.03993},
	urldate = {2022-11-22},
	publisher = {arXiv},
	author = {Yang, Weixin and Lyons, Terry and Ni, Hao and Schmid, Cordelia and Jin, Lianwen},
	month = jul,
	year = {2017},
	note = {arXiv:1707.03993 [cs]
version: 1},
}

@inproceedings{barfoot_batch_2014,
	title = {Batch {Continuous}-{Time} {Trajectory} {Estimation} as {Exactly} {Sparse} {Gaussian} {Process} {Regression}},
	isbn = {978-0-9923747-0-9},
	url = {http://www.roboticsproceedings.org/rss10/p01.pdf},
	doi = {10.15607/RSS.2014.X.001},
	language = {en},
	urldate = {2022-11-02},
	booktitle = {Robotics: {Science} and {Systems} {X}},
	publisher = {Robotics: Science and Systems Foundation},
	author = {Barfoot, Tim and Hay Tong, Chi and Sarkka, Simo},
	month = jul,
	year = {2014},
}

@inproceedings{ratliff_chomp_2009,
	address = {Kobe},
	title = {{CHOMP}: {Gradient} optimization techniques for efficient motion planning},
	isbn = {978-1-4244-2788-8},
	shorttitle = {{CHOMP}},
	url = {http://ieeexplore.ieee.org/document/5152817/},
	doi = {10.1109/ROBOT.2009.5152817},
	urldate = {2022-11-02},
	booktitle = {2009 {IEEE} {International} {Conference} on {Robotics} and {Automation}},
	publisher = {IEEE},
	author = {Ratliff, Nathan and Zucker, Matt and Bagnell, J. Andrew and Srinivasa, Siddhartha},
	month = may,
	year = {2009},
	pages = {489--494},
}

@article{chen_integration_1958,
	title = {Integration of {Paths}-{A} {Faithful} {Representation} of {Paths} by {Noncommutative} {Formal} {Power} {Series}},
	volume = {89},
	issn = {0002-9947},
	url = {http://www.jstor.org/stable/1993193},
	doi = {10.2307/1993193},
	number = {2},
	urldate = {2022-11-01},
	journal = {Transactions of the American Mathematical Society},
	author = {Chen, Kuo-Tsai},
	year = {1958},
	note = {Publisher: American Mathematical Society},
	pages = {395--407},
}

@inproceedings{lyons_rough_2014,
	address = {Korea},
	title = {Rough paths, {Signatures} and the modelling of functions on streams},
	url = {http://arxiv.org/abs/1405.4537},
	language = {en},
	urldate = {2022-07-13},
	booktitle = {Proceedings of the {International} {Congress} of {Mathematicians}},
	author = {Lyons, Terry},
	month = may,
	year = {2014},
}

@article{chen_iterated_1954,
	title = {Iterated {Integrals} and {Exponential} {Homomorphisms}},
	volume = {s3-4},
	issn = {00246115},
	url = {http://doi.wiley.com/10.1112/plms/s3-4.1.502},
	doi = {10.1112/plms/s3-4.1.502},
	language = {en},
	number = {1},
	urldate = {2022-11-01},
	journal = {Proceedings of the London Mathematical Society},
	author = {Chen, Kuo-Tsai},
	year = {1954},
	pages = {502--512},
}

@article{amendola_varieties_2019,
	title = {Varieties of {Signature} {Tensors}},
	volume = {7},
	issn = {2050-5094},
	url = {https://www.cambridge.org/core/product/identifier/S2050509419000033/type/journal_article},
	doi = {10.1017/fms.2019.3},
	language = {en},
	urldate = {2022-11-01},
	journal = {Forum of Mathematics, Sigma},
	author = {Améndola, Carlos and Friz, Peter and Sturmfels, Bernd},
	year = {2019},
	pages = {e10},
}

@article{fermanian_embedding_2021,
	title = {Embedding and learning with signatures},
	volume = {157},
	url = {http://arxiv.org/abs/1911.13211},
	journal = {Comput. Stat. Data Anal.},
	author = {Fermanian, Adeline},
	year = {2021},
	pages = {107148},
}

@article{chen_iterated_1977,
	title = {Iterated path integrals},
	volume = {83},
	journal = {Bulletin of the American Mathematical Society},
	author = {Chen, Kuo-Tsai},
	year = {1977},
	pages = {831--879},
}

@article{boedihardjo_signature_2016,
	title = {The signature of a rough path: {Uniqueness}},
	volume = {293},
	issn = {0001-8708},
	shorttitle = {The signature of a rough path},
	url = {https://www.sciencedirect.com/science/article/pii/S0001870816301104},
	doi = {10.1016/j.aim.2016.02.011},
	language = {en},
	urldate = {2022-11-01},
	journal = {Advances in Mathematics},
	author = {Boedihardjo, Horatio and Geng, Xi and Lyons, Terry and Yang, Danyu},
	month = apr,
	year = {2016},
	pages = {720--737},
}

@article{hambly_uniqueness_2010,
	title = {Uniqueness for the signature of a path of bounded variation and the reduced path group},
	volume = {171},
	issn = {0003-486X},
	url = {http://annals.math.princeton.edu/2010/171-1/p02},
	doi = {10.4007/annals.2010.171.109},
	language = {en},
	number = {1},
	urldate = {2022-11-01},
	journal = {Annals of Mathematics},
	author = {Hambly, Ben and Lyons, Terry},
	month = mar,
	year = {2010},
	pages = {109--167},
}

@inproceedings{king_pregrasp_2013,
	title = {Pregrasp {Manipulation} as {Trajectory} {Optimization}},
	isbn = {978-981-07-3937-9},
	url = {http://www.roboticsproceedings.org/rss09/p15.pdf},
	doi = {10.15607/RSS.2013.IX.015},
	language = {en},
	urldate = {2022-10-25},
	booktitle = {Robotics: {Science} and {Systems} {IX}},
	publisher = {Robotics: Science and Systems Foundation},
	author = {King, Jennifer and Klingensmith, Matthew and Dellin, Christopher and Dogar, Mehmet and Velagapudi, Prasanna and Pollard, Nancy and Srinivasa, Siddhartha},
	month = jun,
	year = {2013},
}

@article{zucker_chomp_2013,
	title = {{CHOMP}: {Covariant} {Hamiltonian} optimization for motion planning},
	volume = {32},
	issn = {0278-3649},
	shorttitle = {{CHOMP}},
	url = {https://doi.org/10.1177/0278364913488805},
	doi = {10.1177/0278364913488805},
	language = {en},
	number = {9-10},
	urldate = {2022-10-11},
	journal = {The International Journal of Robotics Research},
	author = {Zucker, Matt and Ratliff, Nathan and Dragan, Anca D. and Pivtoraiko, Mihail and Klingensmith, Matthew and Dellin, Christopher M. and Bagnell, J. Andrew and Srinivasa, Siddhartha S.},
	month = aug,
	year = {2013},
	note = {Publisher: SAGE Publications Ltd STM},
	pages = {1164--1193},
}

@inproceedings{mukadam_simultaneous_2017,
	title = {Simultaneous {Trajectory} {Estimation} and {Planning} via {Probabilistic} {Inference}},
	isbn = {978-0-9923747-3-0},
	url = {http://www.roboticsproceedings.org/rss13/p25.pdf},
	doi = {10.15607/RSS.2017.XIII.025},
	language = {en},
	urldate = {2022-10-11},
	booktitle = {Robotics: {Science} and {Systems} {XIII}},
	publisher = {Robotics: Science and Systems Foundation},
	author = {Mukadam, Mustafa and Dong, Jing and Dellaert, Frank and Boots, Byron},
	month = jul,
	year = {2017},
}

@misc{lambert_entropy_2021,
	title = {Entropy {Regularized} {Motion} {Planning} via {Stein} {Variational} {Inference}},
	url = {http://arxiv.org/abs/2107.05146},
	urldate = {2022-10-11},
	publisher = {arXiv},
	author = {Lambert, Alexander and Boots, Byron},
	month = jul,
	year = {2021},
	note = {arXiv:2107.05146 [cs]},
}

@article{salvi_signature_2021,
	title = {The {Signature} {Kernel} {Is} the {Solution} of a {Goursat} {PDE}},
	volume = {3},
	issn = {2577-0187},
	url = {https://epubs.siam.org/doi/10.1137/20M1366794},
	doi = {10.1137/20M1366794},
	language = {en},
	number = {3},
	urldate = {2022-07-13},
	journal = {SIAM Journal on Mathematics of Data Science},
	author = {Salvi, Cristopher and Cass, Thomas and Foster, James and Lyons, Terry and Yang, Weixin},
	month = jan,
	year = {2021},
	pages = {873--899},
}

@article{chevyrev_primer_2016,
	title = {A {Primer} on the {Signature} {Method} in {Machine} {Learning}},
	url = {http://arxiv.org/abs/1603.03788},
	urldate = {2022-02-10},
	journal = {arXiv:1603.03788 [cs, stat]},
	author = {Chevyrev, Ilya and Kormilitzin, Andrey},
	month = mar,
	year = {2016},
	note = {arXiv: 1603.03788},
}

@article{chamzas_motionbenchmaker_2022,
	title = {{MotionBenchMaker}: {A} {Tool} to {Generate} and {Benchmark} {Motion} {Planning} {Datasets}},
	volume = {7},
	issn = {2377-3766, 2377-3774},
	shorttitle = {{MotionBenchMaker}},
	url = {https://ieeexplore.ieee.org/document/9645379/},
	doi = {10.1109/LRA.2021.3133603},
	language = {en},
	number = {2},
	urldate = {2022-06-25},
	journal = {IEEE Robotics and Automation Letters},
	author = {Chamzas, Constantinos and Quintero-Pena, Carlos and Kingston, Zachary and Orthey, Andreas and Rakita, Daniel and Gleicher, Michael and Toussaint, Marc and Kavraki, Lydia E.},
	month = apr,
	year = {2022},
	pages = {882--889},
}

@inproceedings{liu_stein_2016,
	title = {Stein variational gradient descent: {A} general purpose bayesian inference algorithm},
	volume = {29},
	url = {https://proceedings.neurips.cc/paper/2016/file/b3ba8f1bee1238a2f37603d90b58898d-Paper.pdf},
	booktitle = {Advances in neural information processing systems},
	publisher = {Curran Associates, Inc.},
	author = {Liu, Qiang and Wang, Dilin},
	editor = {Lee, D. and Sugiyama, M. and Luxburg, U. and Guyon, I. and Garnett, R.},
	year = {2016},
}

@article{zhuo_message_2018,
	title = {Message {Passing} {Stein} {Variational} {Gradient} {Descent}},
	language = {en},
	journal = {Proceedings of the 35th International Conference on Machine Learning},
	author = {Zhuo, Jingwei and Liu, Chang and Shi, Jiaxin and Zhu, Jun and Chen, Ning and Zhang, Bo},
	year = {2018},
	pages = {10},
}

@book{silverman_density_1986,
	address = {Boston, MA},
	title = {Density {Estimation} for {Statistics} and {Data} {Analysis}},
	isbn = {978-0-412-24620-3 978-1-4899-3324-9},
	url = {http://link.springer.com/10.1007/978-1-4899-3324-9},
	language = {en},
	urldate = {2021-01-15},
	publisher = {Springer US},
	author = {Silverman, B. W.},
	year = {1986},
	doi = {10.1007/978-1-4899-3324-9},
}

@inproceedings{lambert_stein_2020,
	title = {Stein {Variational} {Model} {Predictive} {Control}},
	booktitle = {Proceedings of the 4th {Annual} {Conference} on {Robot} {Learning}},
	author = {Lambert, Alexander and Fishman, Adam and Fox, Dieter and Boots, Byron and Ramos, Fabio},
	year = {2020},
}

@inproceedings{barcelos_disco_2020,
	address = {Paris, France},
	title = {{DISCO}: {Double} {Likelihood}-{Free} {Inference} {Stochastic} {Control}},
	copyright = {All rights reserved},
	doi = {978-1-7281-7395-5/20},
	language = {en},
	booktitle = {Proceedings of the 2020 {IEEE} {International} {Conference} on {Robotics} and {Automation}},
	publisher = {IEEE Robotics and Automation Society},
	author = {Barcelos, Lucas and Oliveira, Rafael and Possas, Rafael and Ott, Lionel and Ramos, Fabio},
	month = may,
	year = {2020},
	pages = {7},
}

@article{levine_reinforcement_2018,
	title = {Reinforcement {Learning} and {Control} as {Probabilistic} {Inference}: {Tutorial} and {Review}},
	shorttitle = {Reinforcement {Learning} and {Control} as {Probabilistic} {Inference}},
	url = {http://arxiv.org/abs/1805.00909},
	urldate = {2019-10-31},
	journal = {arXiv:1805.00909 [cs, stat]},
	author = {Levine, Sergey},
	month = may,
	year = {2018},
	note = {arXiv: 1805.00909},
}

@inproceedings{kalakrishnan_stomp:_2011,
	title = {{STOMP}: {Stochastic} trajectory optimization for motion planning},
	shorttitle = {{STOMP}},
	doi = {10.1109/ICRA.2011.5980280},
	booktitle = {2011 {IEEE} {International} {Conference} on {Robotics} and {Automation}},
	author = {Kalakrishnan, M. and Chitta, S. and Theodorou, E. and Pastor, P. and Schaal, S.},
	month = may,
	year = {2011},
	pages = {4569--4574},
}

@inproceedings{williams_aggressive_2016,
	address = {Stockholm},
	title = {Aggressive driving with model predictive path integral control},
	isbn = {978-1-4673-8026-3},
	url = {https://ieeexplore.ieee.org/document/7487277/},
	doi = {10.1109/ICRA.2016.7487277},
	language = {en},
	urldate = {2019-05-08},
	booktitle = {2016 {IEEE} {International} {Conference} on {Robotics} and {Automation} ({ICRA})},
	publisher = {IEEE},
	author = {Williams, Grady and Drews, Paul and Goldfain, Brian and Rehg, James M. and Theodorou, Evangelos},
	month = may,
	year = {2016},
	pages = {1433--1440},
}

@article{blei_variational_2017,
	title = {Variational {Inference}: {A} {Review} for {Statisticians}},
	volume = {112},
	issn = {0162-1459, 1537-274X},
	shorttitle = {Variational {Inference}},
	url = {https://www.tandfonline.com/doi/full/10.1080/01621459.2017.1285773},
	doi = {10.1080/01621459.2017.1285773},
	language = {en},
	number = {518},
	urldate = {2019-04-10},
	journal = {Journal of the American Statistical Association},
	author = {Blei, David M. and Kucukelbir, Alp and McAuliffe, Jon D.},
	month = apr,
	year = {2017},
	pages = {859--877},
}

@article{ramos_hilbert_2016,
	title = {Hilbert maps: {Scalable} continuous occupancy mapping with stochastic gradient descent},
	volume = {35},
	issn = {0278-3649},
	url = {http://journals.sagepub.com/doi/10.1177/0278364916684382},
	doi = {10.1177/0278364916684382},
	number = {14},
	urldate = {2017-09-17},
	journal = {The International Journal of Robotics Research},
	author = {Ramos, Fabio and Ott, Lionel},
	month = dec,
	year = {2016},
	note = {Publisher: SAGE PublicationsSage UK: London, England},
	pages = {1717--1730},
}


\newpage{}
\onecolumn
\appendices
\crefalias{section}{appendix}

\section{Experiments Hyper-parameters}\label{app:exp_hyperparams}
In~\cref{tab:exp_hyperparams} we present the relevant hyper-parameters to reproduce the results in the paper.
It is worth mentioning that the terrain in the 2D motion planning is randomly generated and will vary on each simulation.
Another source of randomness arises when using Monte Carlo samples to approximate the gradient of the log posterior distribution.
Furthermore, due to the stochastic nature of the initial placement of the spline knots, results will vary despite using analytic gradients.

\begin{table}[h]
    \centering
    \begin{tabular}{lccc}
        \toprule
        Parameter & 2D Terrain & Point-mass Navigation & Manipulator Benchmark \\
        \midrule
        Initial state, $\stateInit$ & $\bracks{0.25, 0.75}$ & $\bracks{-1.8, -1.8}$ & Problem dependent \\
        Environment maximum velocity & --- & \SI{5}{\meter\per\sec} & --- \\
        Environment maximum acceleration & --- & --- & --- \\
        Number of spline knots, $\splineKnots$ & 4 & --- & 5 \\
        Number of particles, $\partSize$ & 20 & 30 & 20 \\
        Particle prior & Uniform & $\normal[\stateSeq, \vec{1}]$ & Uniform \\
        Number of action samples, $\polSamples$ & --- & 10 & --- \\
        Cost likelihood inverse temperature, $\temperature$ & 1.0 & 1.0 & 1.0 \\
        Control authority, $\ctrlAuth$ & --- & $5^2$ & --- \\
        Control horizon, $\ctrlHorizon$ & --- & 30 & --- \\
        Stationary kernel, $k\parens{\cdot, \cdot}$ & Squared-exponential & Squared-exponential & Squared-exponential \\
        Stationary Kernel bandwidth, $\lengthscale$ & 1.5 & Silverman's rule & 1.5 \\
        Signature kernel bandwidth, $\lengthscale$ & 1.5 & 5.65 & 1.5 \\
        Signature kernel degree, $\sigDegree$ & 4 & 3 & 6 \\
        Optimiser class & Adam & Adam & Adam \\
        Learning rate, $\stepSize$ & \num{5e-2} & 1 & \num{1e-3} \\
        \bottomrule
    \end{tabular}
    \caption{\label{tab:exp_hyperparams}
        Hyper-parameters used in the experiments.
    }
\end{table}

\section{Path Following Example}\label{sec:exp_obst_field}
\begin{figure*}[h]
    \centering
    \includegraphics{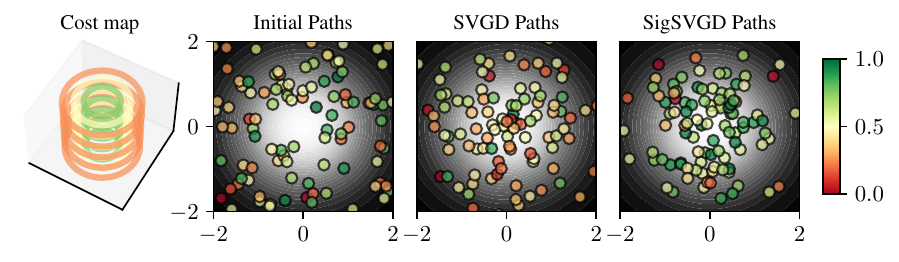}
    \vspace{-10pt}
    \caption{
        \textbf{Qualitative analysis of trajectory tracking task.}
        \emph{Left}: Contour plot of the optimality distribution over sequential time-steps (on $z$-axis).
        \emph{Centre-left}: Cross-section plot at a given time-step of initial path coordinates. The colour of each path indicates its normalised optimality probability.
        \emph{Centre-right}: Cross-section plot of the paths after SVGD optimisation. The sampled paths are diverse and capture the variance of the target distribution. Note, however, that many non-optimal trajectories are close to the origin due to the lack of coordination between consecutive time-steps.
        \emph{Right}: Cross-section plot of the paths after \sigopt\ optimisation. Note how we achieve both diversity and a concentration of optimal paths near the origin.
    }\label{fig:tracking}
\end{figure*}

As a motivating example in~\cref{fig:tracking} we depict the results of a simple two-dimensional path following task.
The goal is to reduce the error between the desired path and candidate paths.
Since we want the error to be as small as possible, the optimal path is one centred at the origin across time.
The objective function is defined as a correlated multivariate Normal distribution across 10 consecutive discrete time-steps such that the optimality likelihood is computed for the entire discretised path.
As the cost function is convex and we are optimising the paths directly---\ie\ not searching for an indirect policy that generates the candidate paths---the solution is trivial.
Nonetheless, the example is useful to illustrate the differences between \sigopt\ and SVMP.

The initial paths are sampled from a uniform distribution and optimised with SVMP and \sigopt\ for 200 iterations.
The length scale of the squared-exponential kernel is computed according to Silverman's rule~\citep{silverman_density_1986} based on the initial sample for \sigopt\ and updated at each iteration using the same method for SVMP. The results in~\cref{fig:tracking} show how both methods are able to promote diversity on the resulting paths.
However, close inspection of the SVMP solution illustrates how coordinates of the candidate paths at each time-step are optimised without coordination, resulting in many paths crisscrossing and non-optimal paths close to the origin.
Conversely, \sigopt\ is promoting diversity of complete paths, rather than coordinates at each cross-sectional time-step, resulting in more direct paths with higher optimality likelihood.

\section{Including Hyper-priors in \sigopt}\label{app:hyperprior}
As mentioned in~\cref{sec:svmp_on_splines}, if one wants to constraint the feasible set of the SVGD optimisation a \emph{hyper-prior} can be included in the algorithm.
Let \( \pdf{h}[\cdot] \) be a hyper-prior and \( \pPdf[\cdot] \) the prior distribution over particles \( \anyvector, \othervector \in \stateSpace \) and recall that the \emph{score function} at each update is computed according to
\begin{equation*}
    {\scoreFunc}^{*} (\anyvector) = \expectation_{\othervector \sim \pPdf} [\big]
    [k^{\oplus} \parens{\othervector, \anyvector} \grad_{\othervector} \log \pPdf[\othervector \given \optimality] + \grad_{\othervector} k^{\oplus} \parens{\othervector, \anyvector}],
\end{equation*}
where the posterior distribution can be factored in $\log \pPdf[\othervector \given \optimality] = \logLik[\optimality \given \othervector] + \log \pPdf[\othervector]$.
We can include the hyper-prior in the formulation by variable substitution.
Let \( \log \pdf{\hat{p}}[\cdot] = \log \pPdf[\cdot] + \log \pdf{h}[\cdot] \), then
\begin{align*}
    {\scoreFunc}^{*} (\anyvector) &= \expectation_{\othervector \sim \pPdf} [\big]
    [k^{\oplus} \parens{\othervector, \anyvector} \grad_{\othervector} \log \pPdf[\othervector \given \optimality] + \grad_{\anyvector} k^{\oplus} \parens{\othervector, \anyvector}] \\
    {\scoreFunc}^{*} (\anyvector) &= \expectation_{\othervector \sim \pPdf} [\bigg]
    [k^{\oplus} \parens{\othervector, \anyvector} \grad_{\othervector} \bracks[\big]{\logLik[\optimality \given \othervector] + \log \pdf{\hat{p}}[\othervector]}
    + \grad_{\othervector} k^{\oplus} \parens{\othervector, \anyvector}] \\
    {\scoreFunc}^{*} (\anyvector) &= \expectation_{\othervector \sim \pPdf} [\bigg]
    [k^{\oplus} \parens{\othervector, \anyvector} \grad_{\othervector} \bracks[\big]{\logLik[\optimality \given \othervector] + \log \pPdf[\othervector] + \log \pdf{h}[\othervector]}
    + \grad_{\othervector} k^{\oplus} \parens{\othervector, \anyvector}], 
\end{align*}
where \( \pdf{h}[\cdot] \) can be any differentiable probability density function. \( \hfill\blacksquare \)

\end{document}